\documentclass[10pt,twocolumn,letterpaper]{article}
\usepackage[accsupp]{axessibility}  
\usepackage{booktabs}       
\usepackage{amsfonts}       
\usepackage{nicefrac}       
\usepackage{microtype}      
\usepackage{xcolor}         
\usepackage{xspace}
\usepackage{titlesec}
\usepackage{subcaption}
\usepackage{titlesec}
\titlespacing*{\section}{0pt}{*0}{*0}
\titlespacing*{\subsection}{0pt}{*0}{*0}
\titlespacing*{\subsubsection}{0pt}{*0}{*0}

\usepackage{wacv}              

\usepackage{multirow} 
\usepackage{makecell}

\newcommand{\blue}[1]{{#1}}

\definecolor{wacvblue}{rgb}{0.21,0.49,0.74}
\usepackage[pagebackref,breaklinks,colorlinks,allcolors=wacvblue]{hyperref}

\def\wacvPaperID{397} 
\def\confName{WACV}
\def\confYear{2026}

\title{LooC: Effective Low-Dimensional Codebook for Compositional Vector Quantization}

\author{Jie Li ~~~~~~~~~~ Kwan-Yee K. Wong ~~~~~~~~~~ Kai Han\thanks{~Corresponding author.}\\
The University of Hong Kong\\
{\tt\small jieli23@hku.hk, kykwong@cs.hku.hk, kaihanx@hku.hk}
}

\begin{document}
\maketitle
\begin{abstract}
Vector quantization (VQ) is a prevalent and fundamental technique that discretizes continuous feature vectors by approximating them using a codebook.
As the diversity and complexity of data and models continue to increase, there is an urgent need for high-capacity, yet more compact VQ methods.
This paper aims to reconcile this conflict by presenting a new approach called \textbf{LooC}, which utilizes an effective \underline{\textbf{L}}ow-dimensional codeb\underline{\textbf{oo}}k for \underline{\textbf{C}}ompositional vector quantization.
Firstly, LooC introduces a parameter-efficient codebook by reframing the relationship between codevectors and feature vectors, significantly expanding its solution space.
Instead of individually matching codevectors with feature vectors, LooC treats them as lower-dimensional compositional units within feature vectors and combines them, resulting in a more compact codebook with improved performance.
Secondly, LooC incorporates a parameter-free extrapolation-by-interpolation mechanism to enhance and smooth features during the VQ process, which allows for better preservation of details and fidelity in feature approximation. 
The design of LooC leads to full codebook usage, effectively utilizing the compact codebook while avoiding the problem of collapse. Thirdly, LooC can serve as a plug-and-play module for existing methods for different downstream tasks based on VQ. 
Finally, extensive evaluations on different tasks, datasets, and architectures demonstrate that LooC outperforms existing VQ methods, achieving state-of-the-art performance with a significantly smaller codebook. 
\end{abstract}
    
\vspace{-0.3cm}
\section{Introduction}
\label{sec:intro}
\vspace{-0.2cm}

Vector quantization (VQ)~\cite{gray1984vector,linde1980algorithm,van2017neural_VQ-VAE} is a widely used technique that converts continuous feature representation into a finite set of discrete vectors, known as the codebook, allowing for efficient analysis and processing for various downstream applications such as 
representation learning~\cite{van2017neural_VQ-VAE,zou2024vqcnir,zheng2025unicode}, 
data compression~\cite{zhu2022unified,williams2020hierarchical,liu2023learning}, clustering~\cite{zhang2009learning,khan2023comprehensive,ribes2011self,hu2023robust}, pattern recognition~\cite{venugopal2018hierarchical,wu2023ridcp,guo2023msmc}, etc.
The VQ codebook is essential for minimizing distortion between input and matched codevectors. This is achieved by clustering features in the latent space to store domain information, resulting in a compact and informative representation.

With the increasing diversity and complexity of data and models, there is an urgent demand for enhanced VQ methods that incorporate efficient and effective codebooks with larger representation capacity.
Increasing the codebook size can potentially improve performance by allowing for a more extensive set of representative codevectors and higher precision in data representation. However, the benefits may plateau while the computational and storage burden continues to grow.  Furthermore, larger codebooks may require more training data to ensure adequate representation, which can be a limiting factor in specific applications.
The trade-off of determining the optimal codebook and codevector sizes involves finding the right balance between accuracy, computational complexity, and storage requirements. It often requires empirical evaluation and experimentation to identify the ideal trade-off for a specific application or dataset.

Many methods have been developed to improve VQ. 
For example,~\cite{lee2022autoregressive_RQ_VAE,zheng2022movq,liu2023learning} combine multiple codebooks to increase the capacity and expressiveness of the codebook for VQ.
\cite{ahmed2005new,li2023resizing} attempt to reduce the size of a learned codebook with a post-processing method while avoiding too much information loss caused by the reduction. \cite{yu2022vector_ViT_VQGAN} introduces a learnable module to quantize lower-dimensional feature vectors by projecting a learned large codebook with high-dimensional codevectors. 
Recent studies~\cite{takida2022_sq_vae,zheng2023online_CVQ} highlight the prevalent issue of codebook collapse in VQ, where only a small subset of codevectors is effectively utilized, limiting the expressive capacity. In response, the state-of-the-art (SOTA) VQ method, CVQ-VAE~\cite{zheng2023online_CVQ}, remedies codebook collapse by updating inactive codevectors using encoded features as anchors, thereby enabling the effective learning of larger codebooks.

Product Quantization (PQ)~\cite{jegou2010product} is a popular method for compressing high-dimensional vectors (such as SIFT descriptors), initially introduced for vector similarity search. 
It can generate an exponentially large codebook at very low memory/time cost. 
PQ introduces the idea of decomposing the space into a Cartesian product of low-dimensional subspaces and quantizing each subspace separately. 
The final codebook of PQ is constructed by taking the Cartesian product of multiple sub-codebooks, with each sub-codebook corresponding to a specific subspace. However, for complex or high-dimensional data, decomposing can generate numerous sub-codebooks, causing the final codebook to become excessively large.
Despite the large codebook size overall, PQ uses separate sub-codebooks for distinct quantizers. This limits the quantization space for each subvector to a single sub-codebook. However, this constraint may hinder PQ's ability to accurately capture fine-grained details of individual subvectors.
To quantize subvectors separately, PQ utilizes multiple distinct quantizers. However, this approach can significantly increase the number of quantizers, especially when decomposing into finer granularity.
Consequently, the complexity in the number of quantizers adds intricacy to the network model and its implementation, which may be considered less elegant.
In this paper, we revisit the idea of decomposing the feature space into compositional sub-spaces and propose a remarkably simple and effective method called LooC, short for \underline{\textbf{L}}ow-dimensional codeb\underline{\textbf{oo}}k for \underline{\textbf{C}}ompositional vector quantization, which introduces several innovative ideas enabling it to not only possess a very compact codebook but also achieve remarkable performance.

\begin{figure}[t]
  \centering
   \includegraphics[width=0.82\linewidth]{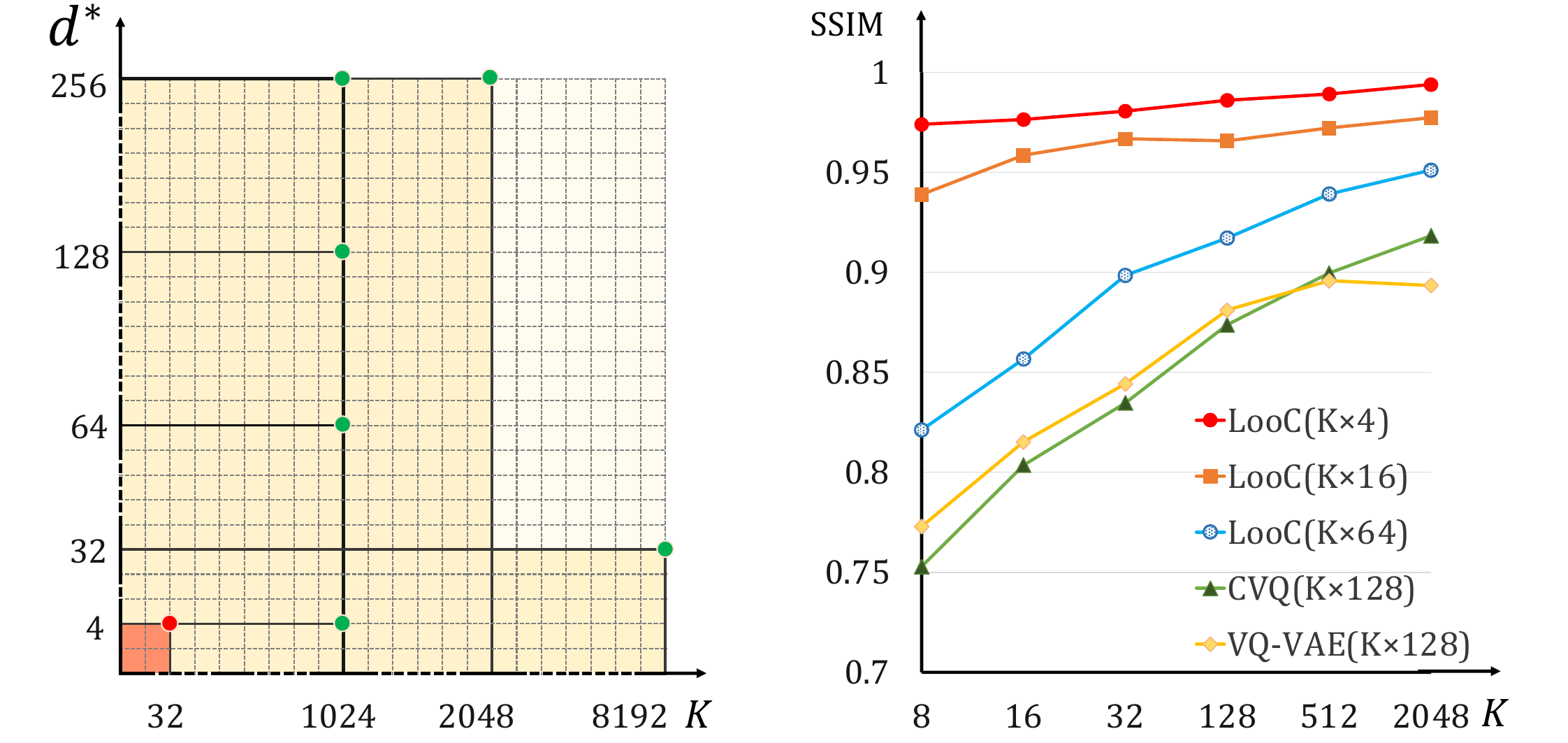}
   \caption{
   \textbf{Codebook size and reconstruction performance.}
   \textbf{Left:} Typical configurations (green dots) of codevector number $K$ and dimension $d^*$ in a codebook. LooC (red dot) stands out with a significantly smaller codebook size of $32\times 4$.
   \textbf{Right:} Reconstruction results on CIFAR10~\cite{krizhevsky2009cifar10}. LooC performs significantly better with a much smaller codebook than other SOTA methods.  
   }
   \label{fig_teaser}
\end{figure}
   
Firstly, we propose a parameter-efficient low-dimensional codebook (LDC), which reframes the relationship between codevectors and feature vectors by considering codevectors as compositional elements within the feature vectors.
This perspective considers the codevectors in LDC as a group of ``visual characters'' that depict the entire visual content rather than individually associating each object feature with a ``visual word''.
Each vector consists of multiple codevectors within a shared codebook, which are obtained using a unified quantizer. 
This design greatly enhances codebook parameterization efficiency, leading to significantly reduced codebook size and notably improved performance.

Secondly, to improve the fidelity of the feature approximation by the codevectors, we further introduce a parameter-free extrapolation-by-interpolation mechanism to enhance and smooth the features during the VQ process to preserve the details better. Notably, before mapping each feature vector to multiple codevectors, we first expand the feature map by interpolation across the spatial dimensions with a factor of $\beta$. The interpolated features are then quantized using our low-dimension codebook, resulting in an extrapolated feature map. 
We obtain the smoothened feature map with the original spatial dimensions by pooling the extrapolated features. 

Thirdly, LooC can be seamlessly integrated as a plug-and-play module to enhance the performance of existing methods for different downstream tasks based on VQ, such as image reconstruction and generation. 
We extensively evaluate LooC on different tasks, datasets, and architectures, obtaining SOTA performance across the board with a much more compact codebook. 
Remarkably, LooC achieves comparable performance on various datasets using a significantly smaller codebook with lower-dimensional codevectors than the SOTA method CVQ~\cite{zhao2022codedvtr}, while maintaining a codebook utilization rate of 100\%. 
For instance, in image reconstruction, LooC utilizes a 1024$\times$ smaller codebook than CVQ to achieve better results. With a much more compact codebook, LooC surpasses the comparative methods in image generation by a large margin, producing highly detailed and realistic images. 

\vspace{-0.2cm}

\section{Related Work}\label{sec:related_work}
\vspace{-0.2cm}

Vector quantization (VQ)~\cite{gray1984vector} is a fundamental research area with origins dating back to the 1980s and remains an enduring subject of interest for many downstream applications.
Classic methods such as LBG~\cite{linde1980algorithm} and Classified-VQ~\cite{cvq1986} use codebooks to represent a set of clustering centers for the input data. 
PQ~\cite{jegou2010product} and OPQ~\cite{ge2013optimized} propose a core paradigm of decomposing high-dimensional features into low-dimensional sub-vectors and quantizing them independently.
Additive Quantization (AQ)~\cite{babenko2014additivequantization} compresses high-dimensional vectors by summing codewords from multiple codebooks without orthogonal subspace decomposition, reducing approximation error and boosting search/classification accuracy while maintaining PQ-like efficiency.
LOPQ~\cite{kalantidis2014locally} partitions high-dimensional data into cells, locally optimizes rotation/space decomposition per cell for residual encoding, with fixed data-size-independent overhead—lower distortion, faster search on billion-scale datasets.
With the rise of deep neural networks, VQ-VAE~\cite{van2017neural_VQ-VAE} introduces VQ into representation learning and employs unsupervised or self-supervised approaches to learn prior knowledge using codebooks. 
VQ-GAN~\cite{esser2021taming_vq_gan} incorporates the vector quantization technique into GAN~\cite{goodfellow2014generative,Radford2015UnsupervisedRL}, leveraging its benefits to enhance the generative model's representation capacity and elevate the quality of sample generation. 
The effectiveness of VQ has led to its widespread adoption in diverse applications across different domains and downstream tasks.
For instance, it has been used in visual recognition \cite{jurie2005creating, zhang2009learning, khan2023comprehensive}, image compression and reconstruction \cite{zhu2022unified, williams2020hierarchical, liu2023learning}, image retrieval \cite{zhu2023revisiting}, image segmentation \cite{kekre2009vq_segmentation, zhao2022codedvtr}, visual and audio generation \cite{dhariwal2020jukebox_vq4music, gal2022image, gu2022vector}, neural radiance field \cite{zhong2023vq_nerf, wallingford2023neural, li2023compact}, knowledge distillation \cite{guo2023predicting}, cross-modal retrieval and translation \cite{cao2017collective, guo2023msmc, lan2023exploring}, vision-language models \cite{dong2023peco}, and other various fields \cite{ribes2011self, ye2012no, venugopal2018hierarchical, lan2023exploring, hu2023robust, wu2023ridcp}.
In these methods, the codebook serves as the prior distribution of the discrete latent space, enabling effective modeling and manipulation of the data distribution.

Significant advancements have been made in optimizing various aspects of the codebook in vector quantization. 
Firstly, to improve codebook's expressiveness and representation capacity, researchers have explored different approaches.
AdaCode~\cite{liu2023learning} learns a set of basis codebooks for each image category and introduces a weight map for adaptive image restoration. MoVQ~\cite{zheng2022movq} incorporates two convolutional layers into the decoder to learn modulation parameters from embedding vectors, converting discrete representations into scaled and biased values.
%
Additionally, efforts have been made to address the codebook collapse problem and increase its usage.
SQ-VAE~\cite{takida2022_sq_vae} introduces stochastic dequantization and quantization techniques to tackle the problem. The SOTA method, CVQ~\cite{zheng2023online_CVQ}, proposes Online Clustered Codebook learning as a solution by updating inactive codevectors using encoded features as anchors.
%
Moreover, various techniques have been proposed to reduce the size of codebooks in VQ.
One approach in~\cite{ahmed2005new} entails sorting the codevectors and then utilizing Huffman coding on the differences between adjacent codevectors.

~\cite{mao2023extreme} adjusts the codebook size using the K-means clustering algorithm. HyperHill~\cite{li2023resizing} utilizes hyperbolic embeddings to enhance codebook vectors with co-occurrence information and rearranges the codebook using the Hilbert curve. Another method~\cite{yu2022vector_ViT_VQGAN} adopts a smaller codebook by introducing a linear projection from the encoder's output to a low-dimensional latent variable space, albeit with an increased number of codebooks.
%
Furthermore, several methods for composite quantizations have been introduced.
SQ~\cite{martinez2014stacked} introduces the approach of iteratively quantizing a vector and its residuals to represent the vector as a stack of codes, known as stacked quantization. RQ~\cite{lee2022autoregressive_RQ_VAE} uses residual to approximate the feature vector recursively in a coarse-to-fine manner. 
\blue{
TQR~\cite{li2021trq} enhances RQ effect for ternary neural networks by combining binarized stem and residual parts.
}

Overall, these techniques consider a feature vector as a unified entity and apply codebooks with the same dimension to process it. In contrast, we consider each feature vector as the composition of multiple smaller codevectors combined through concatenation. 
The effectiveness of segmenting the feature vector into subvectors during quantization has been demonstrated by PQ~\cite{jegou2010product} and SPQ~\cite{jang2021self} in the for visual search.
LooC extends this design to image reconstruction and enhances it to generate embeddings with enhanced representation capabilities.
Unlike PQ, which employs several distinct codebooks, LooC utilizes a single codebook shared across all subvectors to increase the combinatorial nature of all codevectors and thus expand the matching space of the codebook.
PQ constructs the product quantization codebook using handcrafted features and enforces orthogonal constraints. In contrast, LooC directly learns a lightweight and expressive codebook that can faithfully reconstruct the image by training on image reconstruction.
Besides, LooC employs a unified quantizer for all subvectors, unlike PQ which employs multiple distinct sub-quantizers, thus enabling an elegant and unified treatment for all features and codevectors. 

\vspace{-0.2cm}
\section{Method}
\label{sec:method}
\vspace{-0.2cm}

\begin{figure*}[t]
\centering
\includegraphics[width=0.98\linewidth]{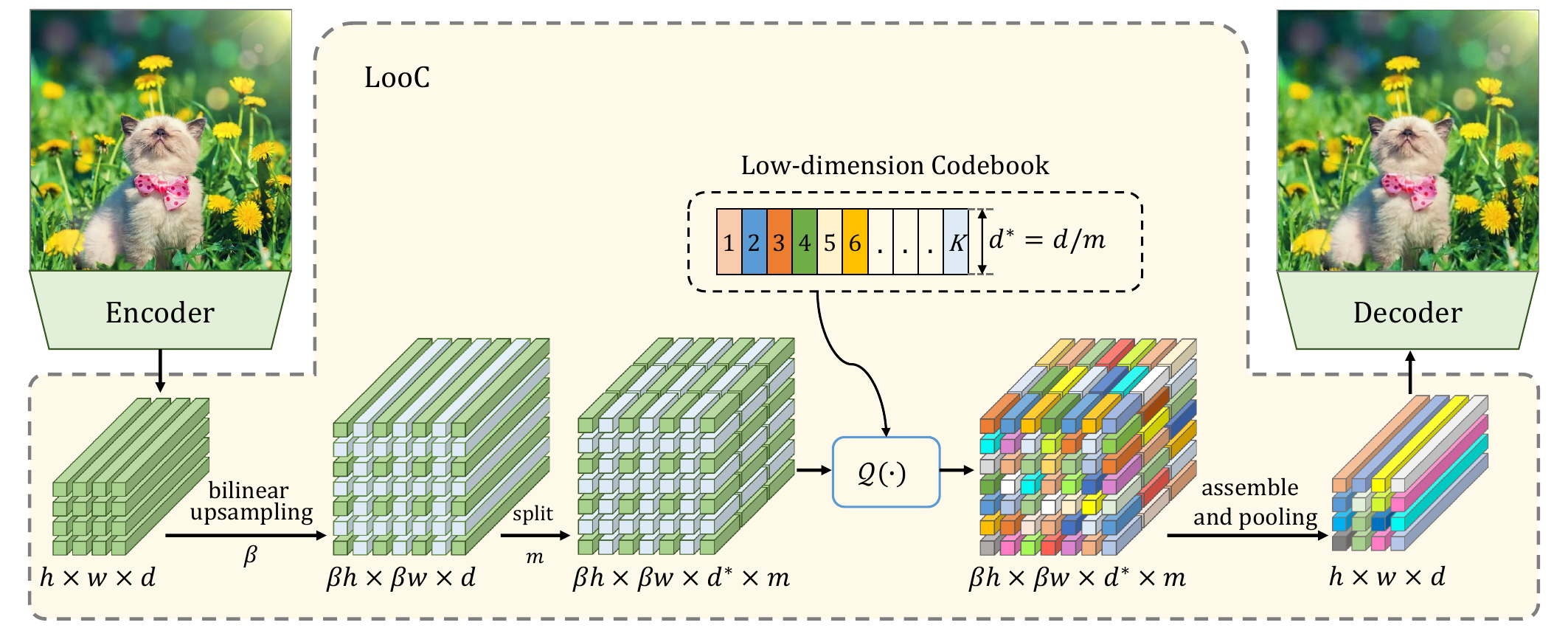}
\caption{
\textbf{Framework of \underline{\textbf{L}}ow-dimensional codeb\underline{\textbf{oo}}k for \underline{\textbf{C}}ompositional vector quantization (LooC).}
The encoder transforms the input image into a continuous latent feature map $z$.
$z$ is then upsampled using bilinear interpolation with scale factor $\beta$. 
Simultaneously, each feature vector in $z$ is divided into $m$ units and quantized using a shared codebook $\mathcal{C}$ containing $K$ codevectors of dimension $d^* = d/m$.
The quantized units are then reassembled and smoothed using average pooling to restore the shape as $z$. 
Finally, the decoder converts the feature map back to the image.
}
\label{fig_pipeline}
\end{figure*}

This section introduces our proposed quantizer, LooC, with an effective low-dimensional codebook for compositional VQ.
We first briefly review VQ-VAE~\cite{van2017neural_VQ-VAE} and PQ~\cite{jegou2010product} as the preliminary.
We then present LooC and elucidate how LooC attains remarkable performance with a low-dimensional codebook and feature enhancement through extrapolation-by-interpolation. 
We also demonstrate how LooC resolves the collapse problem, ensuring the complete utilization of its representation space.

\subsection{Preliminary}
\vspace{-0.2cm}
\paragraph{VQ-VAE for Visual Representation.}
Consider an encoder $\Phi_{\text{ENC}}(\cdot)$ and a decoder $\Phi_{\text{DEC}}(\cdot)$. z
Let $\mathcal{Q}(\cdot)$ denote the quantization operator.
The feature vector $z_{i,j} \in \mathbb{R}^d$ at each spatial location $(i, j)$ is then quantized by replacing it with the most similar codevector in the codebook $\mathcal{C}$, \ie, 
\begin{equation}
\mathcal{Q}(z_{i,j}; \mathcal{C}) = \mathop{\arg\min}\limits_{k \in  [K] }\left \|{z} _{i,j} -  c(k) \right \|,~~~ 
\mathcal{C} = \{(k, c(k))\}_{k=1}^K,
\label{eq:q}
\end{equation}
where $k$ is the code index, $c(k) \in \mathbb{R}^{d^*}$ is the corresponding codevector in $\mathcal{C}$.
To ensure that $z_{i,j}$ can be replaced by $c(k)$, they must have the same dimension, \ie, $d^* = d$.

\vspace{-0.2cm}

\paragraph{Product Quantization (PQ).}
In PQ, an input vector $\bar{z}$ is split into $m$ distinct subvectors $\bar{z}_{\lambda}$, $1 \leq \lambda \leq m$, of dimension $ {d}^{*}=d/m$, where $d$ is a multiple of $m$. The subvectors are quantized separately using $m$ distinct quantizers. The given vector $\bar{z} = (a_1, a_2, \cdots , a_d)$ is therefore mapped as follows:
\begin{equation}
    \begin{matrix}
        \underset{\bar{z}_1(a)}{\underbrace{a_1, \cdots ,  a_{ d^*}} }, 
        \cdots , 
        \underset{\bar{z}_m(a)}{\underbrace{a_ { d - d^* + 1} , \cdots , a_d}  }
        \\
        \rightarrow  q_1 ( \bar{z}_1(a)), \cdots , q_m ( \bar{z}_m (a)),
    \end{matrix}
\label{eq:pq}
\end{equation}
where $q_\lambda$ is a low-complexity quantizer associated with the $\lambda$-th subvector. 
The codebook $\mathcal{C}_\lambda$ can be associated with the corresponding reproduction values $c_{\lambda,k}$, using sub-quantizer $q_\lambda$ .
The complete codebook $\mathcal{C}$ is defined as the Cartesian product of $m$ distinct sub-codebooks $\mathcal{C} = \mathcal{C}_1 \times \cdots \times \mathcal{C}_m$.

\subsection{LooC: Learning low-dimensional Codebook for Compositional VQ}


Instead of employing multiple separate sub-quantizers as in PQ, we adopt a unified quantizer applied to all subvectors in LooC.
Additionally, LooC utilizes a shared codebook for subvectors, eliminating the need for multiple codebooks and preventing the explosive growth of codebooks in PQ by avoiding Cartesian products.
Moreover, LooC incorporates an extrapolation-by-interpolation mechanism that enhances and smooths features, preserving details and ensuring accurate feature approximation.

\vspace{-0.25cm}
\paragraph{Low-dimensional Codebook (LDC).}

Instead of treating codevectors as prototypes like PQ, 
we adopt a different perspective by considering codevectors as compositional units within feature vectors. Analogously, instead of considering the codevectors as ``visual words'' we consider them as ``visual characters''.
By leveraging the inherent compositional nature of feature vectors, LooC effectively captures the underlying structure of the data using a reduced number of codevectors. 
Specifically, for each feature vector $z_{i,j} \in \mathbb{R}^d$, we use a unified quantizer $\mathcal{Q}(\cdot)$ to quantize $m$ sub-vectors at the same time to achieve compositional quantization of $z_{i,j}$ with $m$ codevectors in the codebook
${\mathcal{C}} = \{(k, {c}(k))\}_{k=1}^K$, where $k$ is the code index, ${c}(k)\in \mathbb{R}^{d^*}$ is the corresponding codevector and ${d^*} = {d}/{m}$ is the dimension of each codevector. 
The $m$ code indices can be obtained by quantization operation as follows
$\mathcal{I}_{i,j} = \{\mathcal{Q}(z_{i,j}[(q-1)d^*\colon qd^*]; {\mathcal{C}})\}_{q=1}^m $   
with which $z_{i,j}$ can be quantized by concatenating the codevectors corresponding to $\mathcal{I}_{i,j}$ in ${\mathcal{C}}$.
Diffrent from PQ, we utilize a shared quantizer for all subvectors instead of employing several distinct quantizers. Furthermore, we eliminate the need for multiple sub-codebooks and Cartesian product to obtain the final codebook; instead, a single shared codebook with a compact size is sufficient.

\vspace{-0.25cm}
\paragraph{Feature Enhancement and Smoothness.}
In addition to the low-dimensional codebook design, LooC incorporates a parameter-free extrapolation-by-interpolation mechanism to enhance and smooth features during the vector quantization process, preserving details and ensuring accurate feature approximation.
After obtaining the latent feature map ${z}$ from the encoder $\Phi_{\text{ENC}}(\cdot)$, we first employ bilinear interpolation to interpolate the feature map across the spatial dimensions by a scaling factor $\beta$. This step leads to a larger feature map $z^{it} \in \mathbb{R}^{\beta h\times \beta w \times (d^* \times m)}$ with an increased number of feature vectors for VQ.
Next, for each original feature vector and its interpolated neighbors, we quantize them based on our LDC ${\mathcal{C}}$, leading to an extrapolated feature map $z^{ex} \in \mathbb{R}^{\beta h\times \beta w \times (d^* \times m)}$. This way, the feature expressiveness is enhanced. Finally, we smooth the feature by adopting average pooling on features around each quantized original feature, resulting in a feature map $\Tilde{z} \in \mathbb{R}^{h\times  w \times (d^* \times m)}$, which can then be decoded by the decoder $\Phi_{\text{DEC}}(\cdot)$ to reconstruct the input image. 

\subsection{Codebook Compactness and Exponential Representation Capacity}\label{method_discussion}
A key strength of LDC is that it can produce a large set of ``visual words'' by combining a small set of ``visual characters''.
The feature vector of length $d$ is divided into $m$ sub-vectors, each with a length of $d^* = d/m$.
Thus, the value of $d^*$ determines the granularity of the codevectors in LDC. 
A smaller $d^*$ indicates a more fine-grained representation of the individual components in visual features. Conversely, a larger $d^*$ signifies a coarser level of compositional units.
Using a shared codebook with $K$ codevectors allows for possible combinations of $K^m$ when considering $m$ sub-vectors. 
In contrast, PQ needs multiple independent codebooks to achieve such combinations.
Note that, when $d^* = d$, LDC degenerates to the vanilla codebook. In this case, $m = 1$, which means that each feature vector only has $K$ possible choices from the codebook.

The increase of $m$ leads to the gradual improvement of VQ reconstruction accuracy~\cite{NEURIPS2024_1716d022}. This effect is particularly pronounced when the number of codevectors $K$ is small. As depicted in Fig~\ref{fig_teaser}-Right, we can observe that as $m$ increases, $d^* = d/m$ decreases, and the use of LooC effectively mitigates the adverse impact of reducing $K$. For detailed analysis, see Sec.~\ref{sec:ablation_param_m}.
Note that, in the extreme case where $d^* = 1$, each value in the vector $z_{i,j}$ are all quantized separately. 

The parameter $\beta$ in the extrapolation-by-interpolation operation controls the feature enhancement and smoothness level.
During extrapolation, this operation improves the capacity of its own representation vector by integrating information from neighboring vectors.
Our experiments show that setting $\beta$ to 2 consistently leads to strong performance.


\subsection{Training of LooC}\label{method_train}
An input image $x$ is converted into a feature map $z$ through the encoder $\Phi_{\text{ENC}}(\cdot)$ by $z= \Phi_{\text{ENC}}(x)$.
Let $\hat{z}$ be the quantized feature map of $z$.
The image can then be reconstructed by the decoder with $\hat{x}= \Phi_{\text{DEC}}(\hat{z})$.

The encoder $\Phi_{\text{ENC}}(\cdot)$, decoder $\Phi_{\text{DEC}}(\cdot)$, and codebook $\mathcal{C}$ are jointly optimized by minimizing the loss: 
\begin{equation}
\mathcal{L} = \left \| x - \hat{x}  \right \| _{2}^{2} + \left \| \texttt{sg}\left [  z \right ]- \hat{z}  \right \| _{2}^{2} + \mu  \left \|  z - \texttt{sg}\left [\hat{z}\right ] \right \| _{2}^{2},
\label{eq:loss}
\end{equation}
where $\texttt{sg}$ is a stop gradient operator, the first term is referred to as the reconstruction loss, the second as the codebook loss, and the third as the commitment loss.
We develop our method based on VQ-VAE~\cite{van2017neural_VQ-VAE}.
Moreover, we follow~\cite{zheng2023online_CVQ} to update codevectors to avoid codebook collapse using features as anchors.

\begin{table}[t]
\centering
\scalebox{0.7}
{
\begin{tabular}{lrccccc}\hline
\textbf{Method}  &  & \textbf{~$K\times d^*$} ↓~ 
 & \textbf{~LPIPS ↓} & \textbf{~rFID ↓} & \textbf{~SSIM ↑} & \textbf{~PSNR ↑}  \\ \hline
VQ-VAE~\cite{van2017neural_VQ-VAE} 
& \multirow{5}{*}{\rotatebox{90}{MNIST}}    & $1024 \times 128$ & 0.0282  & 3.43    & 0.9777  & 26.48   \\
HVQ-VAE~\cite{williams2020hierarchical} &   & $1024 \times 128$ & 0.0270  & 3.17    & 0.9790  & 26.90   \\
SQ-VAE~\cite{takida2022_sq_vae}         &   & $1024 \times 128$ & 0.0256  & 3.05    & 0.9819  & 27.49   \\
CVQ-VAE~\cite{zheng2023online_CVQ}      &   & $1024 \times 128$ & 0.0222  & 1.80    & 0.9833  & 27.87   \\
PQ~\cite{jegou2010product}              &   & $256 \times 4 \times $\#$32$ & 0.0120  & 1.76    & 0.9933  & 32.32 \\ 
\textbf{LooC}                           &   & $ 32 \times 4$ &  \textbf{0.0083}  & \textbf{1.70}  & \textbf{0.9961}  & \textbf{35.15} \\
\textbf{LooC}                           &   & $256 \times 4$ &  \textbf{0.0058}  & \textbf{1.31}  & \textbf{0.9976}  & \textbf{37.58} \\ \hline
VQ-VAE~\cite{van2017neural_VQ-VAE}
& \multirow{5}{*}{\rotatebox{90}{CIFAR10}}  & $1024 \times 128$ &  0.2504  & 39.67   & 0.8595  & 23.32   \\
HVQ-VAE~\cite{williams2020hierarchical} &   & $1024 \times 128$ &  0.2553  & 41.08   & 0.8553  & 23.22   \\
SQ-VAE~\cite{takida2022_sq_vae}         &   & $1024 \times 128$ &  0.2333  & 37.92   & 0.8779  & 24.07   \\
CVQ-VAE~\cite{zheng2023online_CVQ}      &   & $1024 \times 128$ &  0.1883  & 24.73   & 0.8978  & 24.72   \\
PQ~\cite{jegou2010product}              &   & $256 \times 4 \times $\#$32$ &  0.0953  & 27.15   & 0.9527  & 28.27   \\ 
\textbf{LooC}                           &   & $ 32 \times 4$ & \textbf{0.0435}  & \textbf{24.53} & \textbf{0.9805}  & \textbf{32.22} \\
\textbf{LooC}                           &   & $256 \times 4$ & \textbf{0.0285}  & \textbf{19.22} & \textbf{0.9880}  & \textbf{34.51} \\ \hline
\end{tabular}
} 
\caption{
\textbf{Image reconstruction results} on low-resolution datasets of MNIST~\cite{lecun1998mnist} and CIFAR10~\cite{krizhevsky2009cifar10}. 
LooC outperforms other SOTA methods with a significantly reduced codebook size of $32 \times 4$, which is $1024 \times$ smaller than  $1024 \times 128$ used by most SOTAs.
}
\label{tab_reconstruction_results}
\end{table}

\begin{table}[t]
\centering
\scalebox{0.8}
{
\setlength\tabcolsep{3pt}
\begin{tabular}{lrccccc}
\hline 
\textbf{Method} 
&  
& \textbf{$K\times d^*$} ↓~  
& \textbf{~Usage↑~} 
& \textbf{~rFID↓}  
& \textbf{~SSIM↑}
& \textbf{~PSNR↑}   \\ \hline
VQGAN~\cite{esser2021taming_vq_gan}
& \multirow{8}{*}{\rotatebox{90}{FFHQ}}     & $ 1024 \times 256  $ & 42\% & 4.42 & 0.6641 & 22.24   \\
ViT-VQGAN~\cite{yu2022vector_ViT_VQGAN}   & & $ 8192 \times ~~32 $ & – & 3.13 & –       & –       \\
RQ-VAE~\cite{lee2022autoregressive_RQ_VAE}& & $ 2048 \times 256  $ & – & 3.88 & 0.6700  & 22.99   \\
MoVQ~\cite{zheng2022movq}                 & & $ 1024 \times ~~64 $ & 56\% & 2.26 & 0.8212  & 26.72   \\
SeQ-GAN~\cite{gu2022rethinking}           & & $ 1024 \times 256  $ & 100\% & 3.12 & –       & –       \\
CVQ-VAE~\cite{zheng2023online_CVQ}        & & $ 1024 \times 256  $ & 100\% & 2.03 & 0.8398  & 26.87   \\
\textbf{LooC-VAE}  &    & \textbf{$ 256  \times ~~~~4$} & 100\% & \textbf{1.97} & \textbf{0.8499} & \textbf{27.73} \\ 
\textbf{LooC-VAE}  &    & \textbf{$ 1024 \times ~~~~4$} & 100\% & \textbf{1.37} & \textbf{0.9276} & \textbf{32.44} \\ \hline
VQGAN~\cite{esser2021taming_vq_gan}   & \multirow{8}{*}{\rotatebox{90}{ImageNet}}  
                    & $ 1024 \times 256  $ & 44\% & 7.94  & 0.5183  & 19.07      \\
ViT-VQGAN~\cite{yu2022vector_ViT_VQGAN}    & & $ 8192 \times ~~32 $  & 96\% & 1.28 & –      & –     \\
RQ-VAE~\cite{lee2022autoregressive_RQ_VAE} & & $ 2048 \times 256  $  & –    & 1.83 & –      & –     \\
MoVQ~\cite{zheng2022movq}                  &  & $ 1024 \times ~~64 $ & 63\% & 1.12 & 0.6731 & 22.42 \\
SeQ-GAN~\cite{gu2022rethinking}            & & $ 1024 \times 256  $  &100\% & 1.99 & –      & –     \\
CVQ-VAE~\cite{zheng2023online_CVQ}         & & $ 1024 \times 256  $  &100\% & 1.57 & 0.7115 & 23.37  \\
\textbf{LooC-VAE} &   & \textbf{$ 256  \times ~~~~4$}   & 100\% & 1.68   & \textbf{0.7233}  & \textbf{23.64} \\
\textbf{LooC-VAE} &   & \textbf{$ 1024 \times ~~~~4$}   & 100\% & \textbf{1.01}   & \textbf{0.7160}  & \textbf{29.15} \\ \hline
\end{tabular}
}
\caption{
\textbf{Image reconstruction} on high-resolution datasets of FFHQ~\cite{karras2019ffhq} and ImageNet~\cite{deng2009imagenet}.
LooC has a compact codebook size of $256 \times 4$, which is $256 \times$ smaller than most SOTA methods'$1024 \times 256$.
}
\label{tab_reconstruction_results_ffhq_imagenet}
\end{table}

\section{Experiments}\label{sec:exp}


\paragraph{Applications.}
To validate the effectiveness of LooC, we conduct experiments in two downstream tasks: \textit{image reconstruction} and \textit{image generation}.
In the reconstruction task, we compare LooC with various VQ methods. For a fair comparison, we integrate LooC into VQ-VAE's training framework by replacing the VQ module, following the SOTA method CVQ~\cite{zheng2023online_CVQ}.
We also reimplement the quantizers in PQ~\cite{jegou2010product} and, like other approaches, apply the VQ-VAE network architecture for training.
Afterward, we assess the generalizability of LooC on larger datasets by employing the VQ-GAN~\cite{esser2021taming_vq_gan} architecture.
Our experiments use the same backbone network and codebook update method as CVQ-VAE~\cite{zheng2023online_CVQ}. 
For image generation, we use the LDM framework~\cite{rombach2022high} and replace the VQ module with LooC and other VQ methods.

\vspace{-0.3cm}
\paragraph{Datasets.} 

We examine and verify our method on various datasets: 
MNIST~\cite{lecun1998mnist}, CIFAR10~\cite{krizhevsky2009cifar10}, and FASHION-MNIST~\cite{xiao2017fashionmnist}. 
After that, we evaluate our method on larger datasets such as ImageNet~\cite{deng2009imagenet}, FFHQ~\cite{karras2019ffhq}, and LSUN~\cite{yu15lsun}. 

\vspace{-0.3cm}
\paragraph{Metrics.}
Following previous works~\cite{zheng2023online_CVQ,van2017neural_VQ-VAE},  
we compare the quality of reconstructed images to their original counterparts using various metrics, including the patch-level structure similarity index (SSIM), feature-level Learned Perceptual Image Patch Similarity (LPIPS)~\cite{zhang2018unreasonable}, image-level Peak Signal-to-Noise Ratio (PSNR), and dataset-level Fréchet Inception Distance (FID)~\cite{heusel2017gans}.

\begin{figure*}[t]
  \centering
   \includegraphics[width=0.98\linewidth]{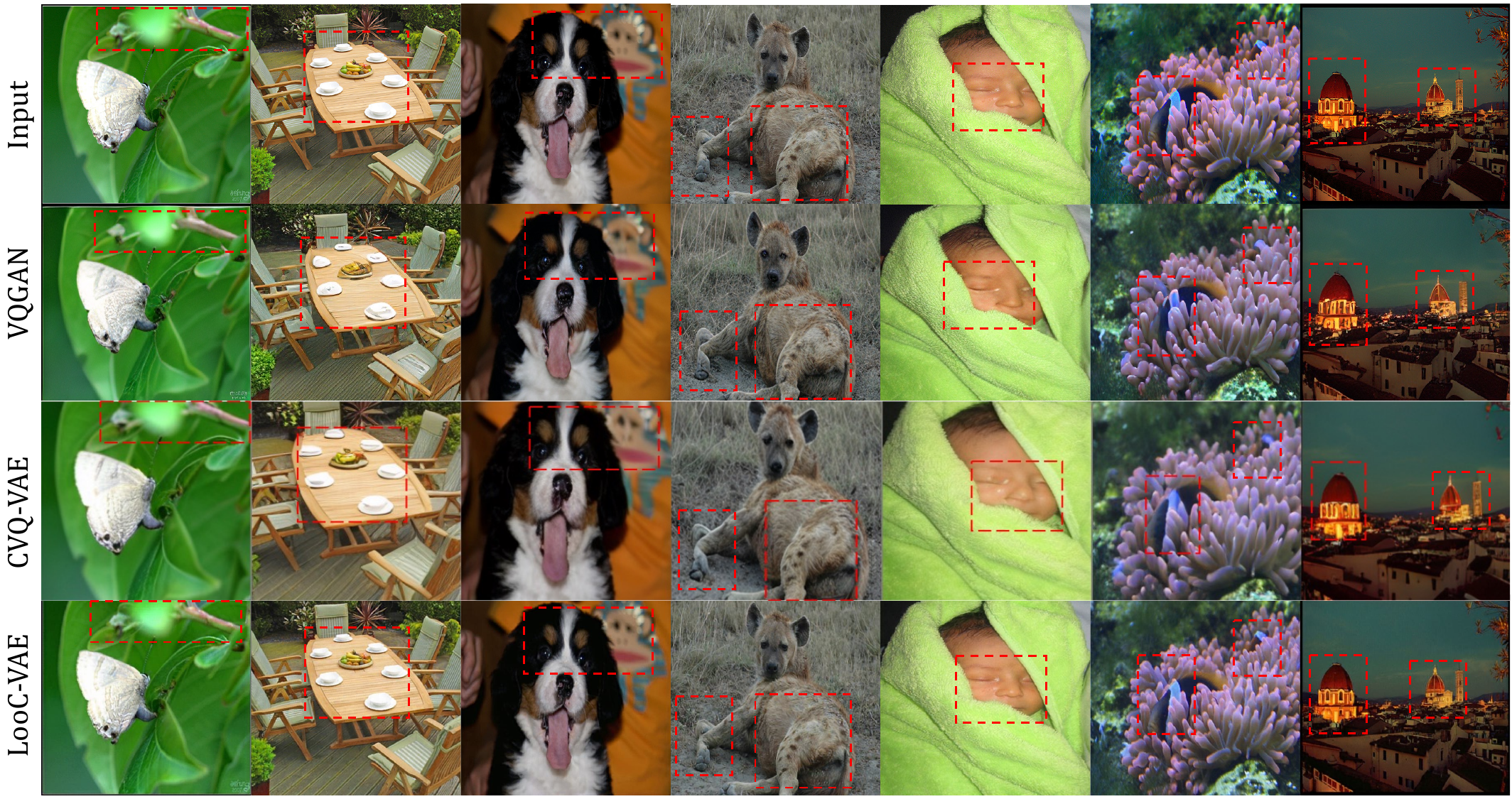}
   \caption{
   \textbf{Qualitative results.}
   Reconstructed images using VQGAN~\cite{esser2021taming_vq_gan}, CVQ~\cite{zheng2023online_CVQ}, and LooC.
   LooC significantly enhances reconstruction quality by preserving image details and restoring texture structures, as highlighted in the red boxes 
   (best viewed in PDF with zoom).
   }
   \label{fig_reconstruction}
\end{figure*}
\subsection{Comparison to Prior Work}
\paragraph{Quantitative Results.}

We conduct experiments on both small datasets like MNIST~\cite{lecun1998mnist} and CIFAR10~\cite{krizhevsky2009cifar10}, as well as large datasets including ImageNet\cite{deng2009imagenet} and FFHQ~\cite{karras2019ffhq}.
In Tab.~\ref{tab_reconstruction_results}, we compare LooC with various VQ methods such as those developed in VQ-VAE~\cite{van2017neural_VQ-VAE}, HVQ~\cite{williams2020hierarchical}, SQVAE~\cite{takida2022_sq_vae}, and CVQ-VAE~\cite{zheng2023online_CVQ}, as well as the reimplemented PQ~\cite{jegou2010product} with 32 independent codebooks.
As shown in Tab.~\ref{tab_reconstruction_results}, it is evident that our method outperforms other techniques across multiple metrics. 
One of the key advantages of our method is that LooC's codebook size is significantly smaller than that of the previous SOTA method CVQ-VAE's. LooC with a size of only $32\times 4$ achieves better performance than CVQ-VAE with a size of $1024\times 128$. In this case, LooC has a $1024\times$ smaller codebook size than CVQ-VAE.
Despite the smaller codebook size, our method achieves significantly better results than CVQ-VAE, as evidenced by the notably reduced rFID score of $1.31$ on MNIST and $19.22$ on CIFAR10.
%
%
Our method also clearly outperforms PQ in all aspects, despite using a significantly smaller codebook. Furthermore, our shared codebook allows us to achieve better results than PQ. The results using the PSNR metric further highlight the strength of our method.

\begin{table*}[t]
\resizebox{1.0\textwidth}{!}
{
\setlength\tabcolsep{4pt}
\begin{tabular}{lllllll}
\hline
\multicolumn{1}{c}{\multirow{2}{*}{\textbf{Method}}} & \multirow{2}{*}{~~$K\times d^*$↓} & \multicolumn{5}{c}{\textbf{MNIST$(28\times28)$ / CIFAR10$(32\times32)$ / FASHION-MNIST$(28\times28)$ }}         \\ 
\cline{3-7} 
\multicolumn{1}{c}{}  & & \multicolumn{1}{c}{\textbf{$l_1$ loss↓}} & \multicolumn{1}{c}{\textbf{LPIPS↓}} & \multicolumn{1}{c}{\textbf{rFID↓}} & \multicolumn{1}{c}{\textbf{SSIM↑}} & \multicolumn{1}{c}{\textbf{PSNR↑}} \\ \hline
VQ-VAE      & $ 1024\times 128  $      & 0.0207 / 0.0527 / 0.0377  & 0.0282 / 0.2504 / 0.0801 & 3.43  / 39.67  / 12.73  & 0.9777 / 0.8595 / 0.9140 & 26.48 / 23.32 / 23.93  \\
CVQ-VAE     & $ 1024\times 128  $      & 0.0180 / 0.0448 / 0.0344  & 0.0222 / 0.1883 / 0.0693 & 1.80  / 24.73  / 8.85   & 0.9833 / 0.8978 / 0.9233 & 27.87 / 24.72 / 24.66  \\ \hline
LooC-VAE        & $256\times ~~4$     & 0.0062 / 0.0144 / 0.0103  & 0.0058 / 0.0285 / 0.0098 & 1.31  / 19.22  / 6.24   & 0.9976 / 0.9880 / 0.9924 & 37.58 / 34.51 / 35.34  \\
LooC-VAE        & $256\times ~~8$     & 0.0068 / 0.0188 / 0.0115  & 0.0064 / 0.0430 / 0.0114 & 1.40  / 24.17  / 6.93   & 0.9972 / 0.9809 / 0.9907 & 36.79 / 32.30 / 34.42  \\
LooC-VAE        & $256\times 16$      & 0.0097 / 0.0246 / 0.0180  & 0.0100 / 0.0681 / 0.0211 & 1.88  / 31.55  / 10.93  & 0.9949 / 0.9679 / 0.9796 & 33.73 / 29.97 / 30.62  \\
LooC-VAE        & $256\times 32$      & 0.0118 / 0.0295 / 0.0236  & 0.0125 / 0.0934 / 0.0325 & 2.19  / 37.58  / 13.95  & 0.9928 / 0.9551 / 0.9660 & 32.05 / 28.43 / 28.20  \\
LooC-VAE        & $256\times 64$      & 0.0138 / 0.0367 / 0.0288  & 0.0157 / 0.1435 / 0.0476 & 2.51  / 51.06  / 17.08  & 0.9903 / 0.9293 / 0.9497 & 30.58 / 26.46 / 26.42  \\ \hline
\end{tabular}

}
\caption{
\textbf{Results under various compositional granularity} using a codebook with $K=256$ codevectors.
As the value of $d^*$ decreases from 64 to 4, achieved by increasing the compositional granularity $m$ from 2 to 32, our method consistently improves performance on all three datasets.
}
\label{tab_ablation_dim}
\end{table*}

\begin{table*}[t]
\resizebox{1.0\textwidth}{!}
{
\setlength\tabcolsep{4pt}
\begin{tabular}{lrlllll}
\hline
\multicolumn{1}{c}{\multirow{2}{*}{\textbf{Method}}} & \multirow{2}{*}{$K\times d^*$~↓~} & \multicolumn{5}{c}{\textbf{MNIST$(28\times28)$ / CIFAR10$(32\times32)$ / FASHION-MNIST$(28\times28)$ }}         \\ 
\cline{3-7} 
\multicolumn{1}{c}{}  & & \multicolumn{1}{c}{\textbf{$l_1$ loss↓}} & \multicolumn{1}{c}{\textbf{LPIPS↓}} & \multicolumn{1}{c}{\textbf{rFID↓}} & \multicolumn{1}{c}{\textbf{SSIM↑}} & \multicolumn{1}{c}{\textbf{PSNR↑}} \\ \hline
VQ-VAE      &$ 1024\times 128  $ & 0.0207 / 0.0527 / 0.0377  & 0.0282  / 0.2504  / 0.0801 & 3.43  / 39.67  / 12.73  & 0.9777  / 0.8595 / 0.9140 & 26.48  / 23.32 / 23.93  \\
CVQ-VAE     &$ 1024\times 128  $ & 0.0180 / 0.0448 / 0.0344  & 0.0222  / 0.1883  / 0.0693 & 1.80  / 24.73  / ~~8.85 & 0.9833  / 0.8978 / 0.9233 & 27.87  / 24.72 / 24.66  \\ \hline
LooC-VAE    & $~~256 \times 128$       & 0.0166 / 0.0464 / 0.0344  & 0.0213  / 0.2144  / 0.0680 & 3.17  / 64.71  / 21.76  & 0.9853  / 0.8923 / 0.9256 & 28.63  / 24.46 / 24.71  \\
LooC-VAE    & $~~512 \times ~~64$        & 0.0122 / 0.0343 / 0.0266  & 0.0133  / 0.1249  / 0.0410 & 2.28  / 45.23  / 16.02  & 0.9923  / 0.9392 / 0.9566 & 31.77  / 27.09 / 27.16  \\
LooC-VAE    & $1024 \times ~~32$       & 0.0094 / 0.0261 / 0.0197  & 0.0095  / 0.0735  / 0.0243 & 1.84  / 33.51  / 11.59  & 0.9952  / 0.9643 / 0.9760 & 34.10  / 29.45 / 29.83  \\
LooC-VAE    & $2048 \times ~~16$       & 0.0073 / 0.0205 / 0.0138  & 0.0070  / 0.0490  / 0.0144 & 1.51  / 26.29  / ~~8.29  & 0.9968  / 0.9773 / 0.9873 & 36.14  / 31.55 / 32.84  \\
LooC-VAE    & $4096 \times ~~~~8$        & 0.0056 / 0.0146 / 0.0098  & 0.0049  / 0.0293  / 0.0091 & 1.17  / 19.40  / ~~5.91  & 0.9980  / 0.9878 / 0.9929 & 38.39  / 34.40 / 35.71  \\
LooC-VAE    & $8192 \times ~~~~4$        & 0.0047 / 0.0099 / 0.0083  & 0.0040  / 0.0154  / 0.0070 & 0.99  / 12.65  / ~~4.86  & 0.9984  / 0.9937 / 0.9947 & 39.71  / 37.49 / 37.05  \\ \hline
\end{tabular}
}
\caption{
\textbf{Results under a fixed codebook size of $s=K \times d^*$ with varying $K$ and $d^*$.}
The compositional granularity parameter $m$ changes proportionally to $d^*$ as $m = d/d^*$. 
}
\label{tab_ablation_keep_same_size}
\end{table*}

\begin{table*}[t]
\centering
\resizebox{1.0\textwidth}{!}
{

\setlength\tabcolsep{4pt}
\begin{tabular}{lrlllll}
\hline
\multicolumn{1}{c}{\multirow{2}{*}{\textbf{Method}}} & \multirow{2}{*}{$K\times d^*$~↓~} & \multicolumn{5}{c}{\textbf{MNIST$(28\times28)$ / CIFAR10$(32\times32)$ / FASHION-MNIST$(28\times28)$ }}         \\ 
\cline{3-7} 
\multicolumn{1}{c}{}  & & \multicolumn{1}{c}{\textbf{$l_1$ loss↓}} & \multicolumn{1}{c}{\textbf{LPIPS↓}} & \multicolumn{1}{c}{\textbf{rFID↓}} & \multicolumn{1}{c}{\textbf{SSIM↑}} & \multicolumn{1}{c}{\textbf{PSNR↑}} \\ \hline

VQ-VAE    &$ 1024\times 128  $&  0.0207 / 0.0527 / 0.0377  &  0.0282 / 0.2504 / 0.0801  &  3.43 / 39.67 /  12.73  &  0.9777 / 0.8595 / 0.9140  &  26.48 / 23.32 / 23.93 \\
CVQ-VAE   &$ 1024\times 128  $&  0.0180 / 0.0448 / 0.0344  &  0.0222 / 0.1883 / 0.0693  &  1.80 / 24.73 / ~~8.85  &  0.9833 / 0.8978 / 0.9233  &  27.87 / 24.72 / 24.66 \\ \hline
LooC-VAE      & $ ~~~~~~8 \times 4$ &  0.0095 / 0.0220 / 0.0168  &  0.0097 / 0.0559 / 0.0197  &  1.86 / 28.35 / ~~9.92  &  0.9950 / 0.9740 / 0.9819  &  33.90 / 30.96 / 31.26 \\
LooC-VAE      & $  ~~~~16 \times 4$ &  0.0079 / 0.0211 / 0.0162  &  0.0080 / 0.0521 / 0.0184  &  1.61 / 27.13 / ~~9.89  &  0.9964 / 0.9763 / 0.9833  &  35.53 / 31.32 / 31.53 \\
LooC-VAE      & $  ~~~~32 \times 4$ &  0.0082 / 0.0189 / 0.0145  &  0.0083 / 0.0435 / 0.0158  &  1.70 / 24.53 / ~~8.73  &  0.9961 / 0.9805 / 0.9864  &  35.15 / 32.22 / 32.49 \\
LooC-VAE      & $  ~~~~64 \times 4$ &  0.0075 / 0.0178 / 0.0133  &  0.0072 / 0.0407 / 0.0140  &  1.54 / 23.70 / ~~8.05  &  0.9967 / 0.9825 / 0.9883  &  36.02 / 32.73 / 33.26 \\
LooC-VAE      & $   ~~128 \times 4$ &  0.0068 / 0.0158 / 0.0114  &  0.0065 / 0.0332 / 0.0112  &  1.43 / 21.15 / ~~6.90  &  0.9972 / 0.9859 / 0.9910  &  36.77 / 33.70 / 34.59 \\
LooC-VAE      & $   ~~256 \times 4$ &  0.0062 / 0.0144 / 0.0103  &  0.0058 / 0.0285 / 0.0098  &  1.31 / 19.22 / ~~6.24  &  0.9976 / 0.9880 / 0.9924  &  37.58 / 34.51 / 35.34 \\
LooC-VAE     & $   ~~512 \times 4$ &  0.0055 / 0.0137 / 0.0092  &  0.0051 / 0.0258 / 0.0083  &  1.18 / 18.14 / ~~5.50  &  0.9980 / 0.9891 / 0.9937  &  38.40 / 34.92 / 36.24 \\
LooC-VAE      & $    1024 \times 4$ &  0.0051 / 0.0116 / 0.0083  &  0.0045 / 0.0192 / 0.0073  &  1.10 / 14.10 / ~~5.07  &  0.9982 / 0.9918 / 0.9946  &  38.97 / 36.38 / 37.10 \\
\hline
\end{tabular}

}
\caption{
\textbf{Results with various $K \in \{8, 16, \cdots, 1024 \}$} of codevectors with fixed $d^* = 4$. 
}
\label{tab_ablation_num}
\end{table*}

Next,  in Tab.\ref{tab_reconstruction_results_ffhq_imagenet}, we verify the effectiveness of our method on more challenging large-scale datasets with high resolution, FFHQ~\cite{karras2019ffhq} and ImageNet~\cite{deng2009imagenet}.
Our LooC-VAE is compared to current SOTA methods, including CVQ-VAE~\cite{zheng2023online_CVQ}, VQ-GAN~\cite{esser2021taming_vq_gan}, ViT-VQGAN~\cite{yu2022vector_ViT_VQGAN}, RQ-VAE~\cite{lee2022autoregressive_RQ_VAE}, SeQ-GAN~\cite{gu2022rethinking}, and MoVQ~\cite{zheng2022movq}, for the reconstruction task.
Our method consistently outperforms previous SOTA methods across both datasets, as revealed in Tab.~\ref{tab_reconstruction_results_ffhq_imagenet}. In comparison to CVQ-VAE~\cite{zheng2023online_CVQ}, our LooC-VAE ($256\times4$) achieves similar or better results while using a $256 \times$ smaller codebook.
 
Simultaneously, our method achieves 100\% codebook utilization as shown in Tab.\ref{tab_reconstruction_results_ffhq_imagenet}. This enables the optimization of all codevectors and ensures the parameter efficiency of the codebook.
These experimental results demonstrate that our codebook has significant advantages in being lightweight in terms of dimension and quantity of codevectors. 

\vspace{-0.4cm}
\paragraph{Qualitative Results.}
In Fig.~\ref{fig_reconstruction}, we compare the visualization results between our method, LooC-VAE, and the SOTA techniques, including VQ-GAN~\cite{esser2021taming_vq_gan} and CVQ-VAE~\cite{zheng2023online_CVQ}. 
Our method excels in preserving image details and restoring texture structures, offering significant advantages over other methods.
This is evident when examining the regions highlighted by the red boxes in Fig~\ref{fig_reconstruction}. 
Notably, the \textit{fruits on the table} in the second column, the \textit{paws of the hyena} in the fourth column, and the \textit{buildings} in the last column demonstrate the superiority of our approach. 
Our approach achieves high fidelity in restoring the original image's appearance, unlike other methods that may cause losses or distortions. This is crucial for downstream tasks.

\vspace{-0.2cm}
\subsection{Expressiveness of LooC}
\vspace{-0.1cm}
In this section, we investigate the reasons behind LooC's exceptional performance and compact design. We primarily focus on the compositional VQ and extrapolation-by-interpolation operations, which are the two most critical components of LooC.

\vspace{-0.3cm}
\paragraph{Enhanced Capacity via Fine-Grained Combination.}
\label{sec:ablation_param_m}
Analyzing Fig.~\ref{fig_teaser}-Right, we make two notable observations during our preliminary performance analysis of LooC. Firstly, as the number of codevectors $K$ increases, the performance of all methods shows improvement. 
Secondly, our approach surpasses other SOTA methods, exhibiting a more substantial advantage specifically when $K$ is smaller (\eg, $K=32$). Additionally, our method demonstrates improved performance as $m$ increases, leading to a lower dimension $d^*$ in our LooC.

Two key findings emerge from this analysis. 
Firstly, the increase in codebook size by adopting a larger $K$ substantially impacts VQ's accuracy.
This is consistent with the fact that existing methods exploit larger $K$ to enhance the effectiveness of VQ, highlighting the challenging nature of reducing codebook size.
Secondly, our LooC exhibits greater adaptability when $K$ is small, while maintaining excellent accuracy performance. LooC achieves this by controlling the granularity parameter $m$ in compositional VQ, with larger $m$ values demonstrating better adaptability to small $K$ values.

We further verify the importance of the compositional VQ through experiments, which are shown in Tab.~\ref{tab_ablation_dim}.
In the experiment, we set $K=256$ and vary the value of $m$, gradually increasing it from $2$ to $32$. 
The dimension $d$ of each latent feature vector is $128$, resulting in a decrease of $d^*$ from $64$ to $4$.
Our method demonstrates consistent performance improvement across three different datasets. For instance, on CIFAR10, the PSNR gradually increases from $26.46$ to $34.51$, and the rFID decreases from $51.06$ to $19.22$. Similarly, other indicators also exhibit a gradual and consistent improvement. This highlights the effectiveness of our approach, which exploits fine-grained combinations, enhancing the capacity of the codebook and leading to better overall performance.


\vspace{-0.3cm}
\paragraph{Compact Codebook with Low-dimension.}
\label{sec_analyse_k}
\begin{figure*}[t]
\centering
\includegraphics[width=0.98\linewidth]{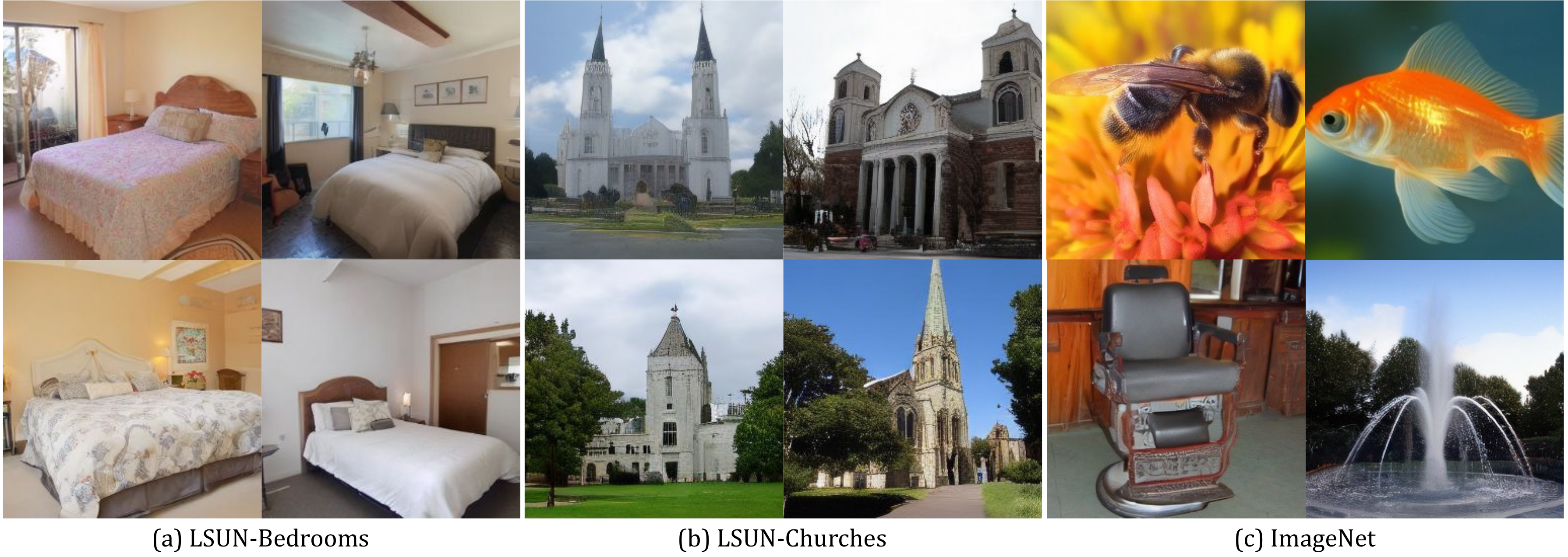}

\caption{
\textbf{Unconditional image generation} on LSUN~\cite{yu15lsun} and \textbf{class-conditional image generation} on Imagenet~\cite{deng2009imagenet}.
}
\label{fig_generation}
\end{figure*}

In Tab.~\ref{tab_ablation_keep_same_size}, we maintain a constant total codebook size of $s = K \times d^*$, while adjusting $K$ and $d^*$. As $d^*$ decreases, resulting in an increase of $m = d / d^*$, $K$ increases as well. This indicates that smaller individual codevectors allow for more codevectors to be added while still maintaining the same overall codebook size.
The experimental results
clearly demonstrate that as $d^*$ decreases ( or $m$ increases), the effectiveness of our method gradually improves, expanding its advantages over existing SOTA approaches.
For instance, by increasing the value of $m$ from 1 to 32 in LooC($K=256, d^* = 128/m$) on the MNIST dataset, we observe a significant improvement in PSNR, increasing from $28.6$ to $39.7$. 
All three datasets showed improved performance in various metrics.
Based on this fact, we investigate the possibility of reducing the number of codevectors in the codebook. The exploration results are shown in Tab.~\ref{tab_ablation_num}.
We kept $m$ fixed at 32 by maintaining $d^* = 4$, and only varied the value of $K$ in our experiments.
Our method is still able to maintain a high 
performance
even when $K$ is small.
Furthermore, the performance of our method improves progressively as $K$ increases.
For instance, when referring to the rFID, our method with a codebook ($K=32, d^*=4$) achieves results similar to CVQ ($K=1024,d^*=128$) on the CIFAR10 dataset. This corresponds to a reduction in codebook size by a factor of $1024\times$. Similar observations are made on the MNIST and FASHION-MNIST datasets.


\begin{table}[t]

\centering
\scalebox{0.88}
{
\setlength\tabcolsep{4pt}

\begin{tabular}{llrllll}
\hline
\multicolumn{1}{c}{\textbf{Method}} & \textbf{$\beta$} & $K \times d^*$~↓~      & \multicolumn{1}{c}{\scalebox{0.88}{\textbf{LPIPS↓}}} & \multicolumn{1}{c}{\scalebox{0.88}{\textbf{rFID↓}}} & \multicolumn{1}{c}{\scalebox{0.88}{\textbf{SSIM↑}}} & \scalebox{0.88}{\textbf{PSNR↑}}  \\ \hline
VQ-VAE                     &  - &$ 1024\times 256  $& 0.1175                    & 4.42                     & 0.6641                   & 22.24 \\
CVQ-VAE                    &  - &$ 1024\times 256  $& 0.0533                    & 2.03                     & 0.8398                   & 26.87 \\
LooC-VAE                       & 2 & $1024\times 256$     & 0.0532                    & 1.83                     & 0.8627                   & 27.02 \\ \hline
LooC-VAE                       & 1 & $~~256\times ~~~~4$  & 0.0528                    & 2.27                     & 0.8571                   & 27.19 \\
LooC-VAE                       & 2 & $~~256\times ~~~~4$  & 0.0501                    & 1.97                     & 0.8499                   & 27.73 \\
LooC-VAE                       & 3 & $~~256\times ~~~~4$  & 0.0478                    & 1.89                     & 0.8523                   & 27.00 \\
LooC-VAE                       & 4 & $~~256\times ~~~~4$  & 0.0490                    & 2.01                     & 0.8510                   & 26.88 \\ \hline
\end{tabular}
}
\caption{\textbf{Results on various settings} on FFHQ~\cite{karras2019ffhq}.
}
\label{tab_ablation_beta}
\end{table}

\vspace{-0.3cm}
\paragraph{Parameter-free Extrapolation-by-interpolation.}
\label{sec:ablation_param_beta}
Tab.~\ref{tab_ablation_beta} evaluates the effectiveness of the extrapolation-by-interpolation operation from two perspectives.
Firstly, we compare our approach with the SOTA method CVQ~\cite{zheng2023online_CVQ} on the high-resolution dataset FFHQ~\cite{karras2019ffhq}. For this comparison, we set $m=1$, making our LDC equivalent to the standard codebook. Other parameters such as $K=1024$ and $d^* = d =256$ remain consistent with CVQ.
We then apply our proposed extrapolation-by-interpolation mechanism with $\beta=2$.
The results show that our proposed mechanism significantly improves VQ's accuracy. For instance, compared to CVQ, our method reduces the rFID score from $2.03$ to $1.83$, while increasing the PSNR from $26.87$ to $27.02$.

Subsequently, we utilize LooC with $K=256$ and $d^* = 4$, setting different $\beta \in \{1,2,3,4\}$. The corresponding experimental results are presented in Tab.~\ref{tab_ablation_beta}. We find that employing a bilinear difference with $\beta=2$ can result in significant improvements. As $\beta$ increases, the spatial correlation of the vectors weakens, leading to a gradual decline in the effectiveness of the improvement. Therefore, we choose $\beta=2$, and accordingly, we can combine this mechanism with other VQ techniques to further improve accuracy.


\vspace{-0.2cm}
\subsection{Plug-and-play for Image Generation}
\vspace{-0.1cm}
We use our quantizer LooC to assess its effectiveness in image generation, following the advanced CVQ~\cite{zheng2023online_CVQ} method.
We replace the VQ module in VQGAN~\cite{esser2021taming_vq_gan} with LooC and then apply it to the LDM~\cite{rombach2022high} method, training it on LSUN~\cite{yu15lsun} and ImageNet~\cite{deng2009imagenet}. The results of unconditional image generation on LSUN and category-conditional image generation on ImageNet are presented in Fig.~\ref{fig_generation}. 
Our approach produces highly detailed and realistic images, demonstrating that our method is a practical plug-and-play module suitable for various downstream tasks.

\vspace{-0.3cm}
\section{Conclusion} \label{sec:conclusion}
\vspace{-0.2cm}
We have presented LooC, a highly efficient quantizer with a low-dimensional codebook for compositional vector quantization. 
LooC not only delivers exceptional performance but also has a remarkably compact codebook.
By treating codevectors as compositional units within feature vectors, LooC achieves a more condensed codebook without compromising performance. Moreover, LooC incorporates an extrapolation-by-interpolation mechanism that enhances and smooths features, ensuring accurate feature approximation and preserving intricate details.
Our quantizer offers a simple yet effective solution that can seamlessly integrate into existing architectures for representation learning. 

\section*{Acknowledgement}
This work is supported by National Natural Science Foundation of China (Grant No. 62306251), Hong Kong Research Grants Council - General Research Fund (Grant No.: 17211024), and HKU Seed Fund for Basic Research.
{
    \small
    \bibliographystyle{ieeenat_fullname}
    \bibliography{main}
}
\clearpage
\clearpage
\setcounter{page}{1}

\setcounter{section}{0}




\twocolumn[
\begin{@twocolumnfalse}
\begin{center}
{\large\bf LooC: Effective Low-Dimensional Codebook for Compositional Vector Quantization
}\\
\vspace{1em}
{\large\bf--- Supplementary Material ---}
\end{center}
\end{@twocolumnfalse}
]


%

\section{Storage and Computational Efficiency}
Sec.~\ref{method_discussion} shows that our codebook theoretically requires only $K'=K^{\frac{1}{m}}$ codevectors to achieve the same capacity as an ordinary VQ with $K$ codevectors. Particularly, when $m>1$, $K'$ is significantly smaller than $K$. 
Here, we delve deeper into analyzing the storage and computing costs associated with LooC.
It shows that LooC provides greater capacity while consuming less space and computation than traditional VQ, making LooC a much more efficient alternative.

\subsection{Storage Efficiency}
\paragraph{Storage Cost of Codebook.}
LooC requires fewer codevectors ($K'$) and lower dimension ($d$) compared to traditional VQ, resulting in less storage cost.
When using a codebook containing $K$ codevectors of dimension $d$ to perform a VQ operation on the feature map ${z} \in \mathbb{R} ^{h\times w\times d}$, the codebook needs to store $K \times d$ values, usually in 32-bit floating point format. 
Therefore, a total of $S_{\text{codebook}} = 32\times K \times d$ bits of storage is required.
However, in our LooC, there are $K'\times d^* = K' \times d / m$ values to be stored. In theory, this only requires $S'_{\text{codebook}} = 32\times K^{\frac{1}{m}} \times d/m$ bits of storage when $K'= K^{\frac{1}{m}}$. As the $K'$ value increases, our method can achieve larger capacity while consuming less storage than traditional methods.

\paragraph{Storage Cost of Indices.}
When performing the VQ operation on the codebook of $K$ codevectors, the $h \times w$ indices are needed to store the corresponding relationship of quantized matching. Each index requires $\log_{2}{K} $ bits of storage.
Therefore, the total required storage is $S_{\text{index}}=h\times w \times \log_{2}{K} $. In our LooC method, since each feature vector is divided into $m$ component units, $h \times w \times m$ indices are required, occupying  $S'_{\text{index}}=h\times w \times m \times \log_{2}{K'}$ bits of storage. It is worth noting that $K'<K$. When $K'=K ^ {\frac{1}{m}}$, $S'_{\text{index}}$ is equivalent to $S_{\text{index}}$.

\subsection{Computational Efficiency}
To determine the most similar matches between the $K$ codevectors in the codebook and the feature representation ${z} \in \mathbb{R} ^{h\times w\times d}$, a total of $h \times w \times K$ similarity calculations are needed. Here, we use the widely adopted cosine similarity for matching purposes. Consequently, in conventional VQ,  $h \times w \times K \times d$ multiplication operations are needed, while omitting the addition operations. 
However, in our LooC with $K'$ codevectors, the feature representation $z$ is decomposed into $m$ segments, leading to the need for $ h \times w \times m \times K' \times d^*$ multiplication operations. 
Here, $d^* = d/m$. Therefore, it requires $h \times w \times m \times K' \times (d/m) = h \times w \times K' \times d$ multiplication operations.
Since $K'$ is much smaller than $K$, the computational cost of our method is also much smaller than that of conventional VQ.

\begin{figure}[b]
\centering
\includegraphics[width=1.0\linewidth]{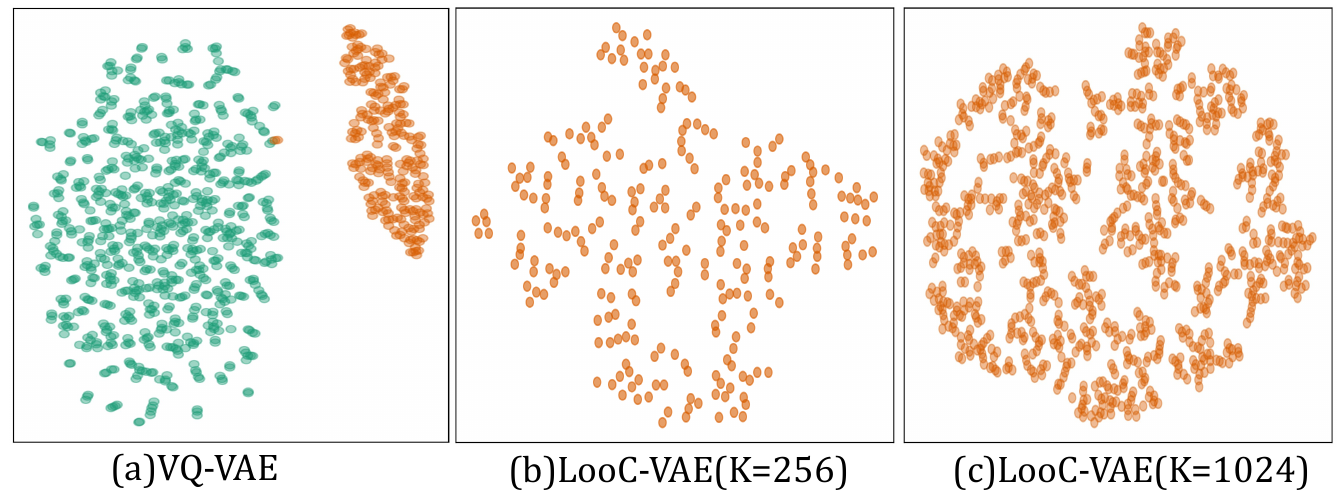}
\caption{
\textbf{Codebook visualization with t-SNE} for models trained on CIFAR10 and evaluated on the validation set. 
VQ-VAE has unused codevectors (green points) with only $24.12\%$ useage. LooC achieves $100\%$ usage at both $K=256$ and $K=1024$.
}
\label{fig_distribution}
\end{figure}

\begin{figure*}[tb]
  \centering
   \includegraphics[width=\textwidth]{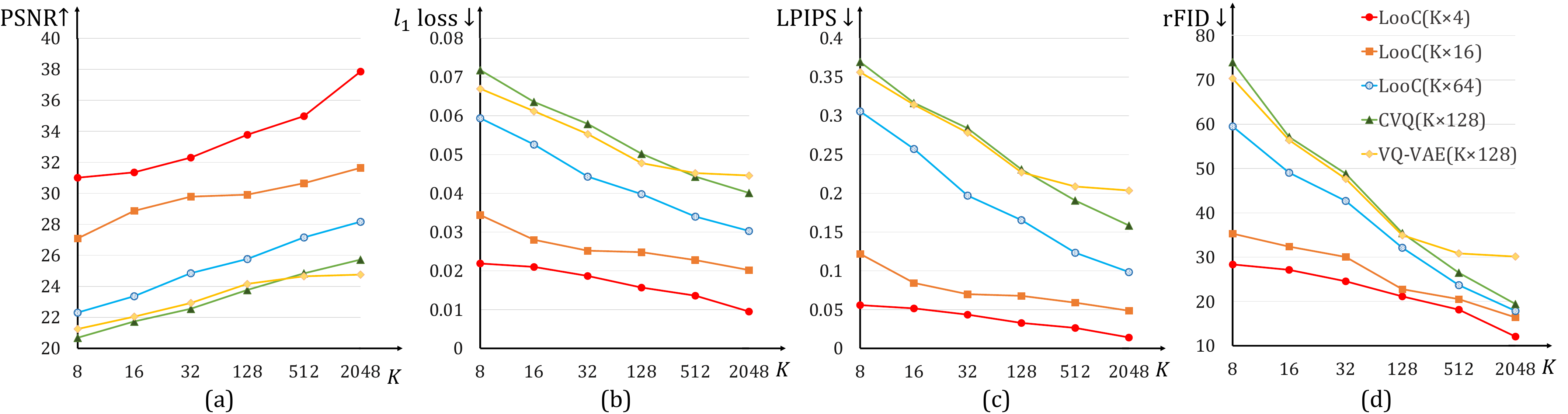}
   \caption{
   \textbf{Reconstruction results} of LooC and other SOTA methods on CIFAR10~\cite{krizhevsky2009cifar10} with various numbers of codevectors demonstrate LooC's superior performance with a smaller codebook, showcasing its flexibility and efficiency. Reducing the codevector dimension $d^*$, \ie, increasing the value of $m = d/d^*$, in LooC leads to a more detailed combinational quantization and improved performance,  especially for small values of $K$.
   }
   \label{fig_teaser_extension}
\end{figure*}
  

\begin{figure}[tb]
\centering
\includegraphics[width=0.7\linewidth]{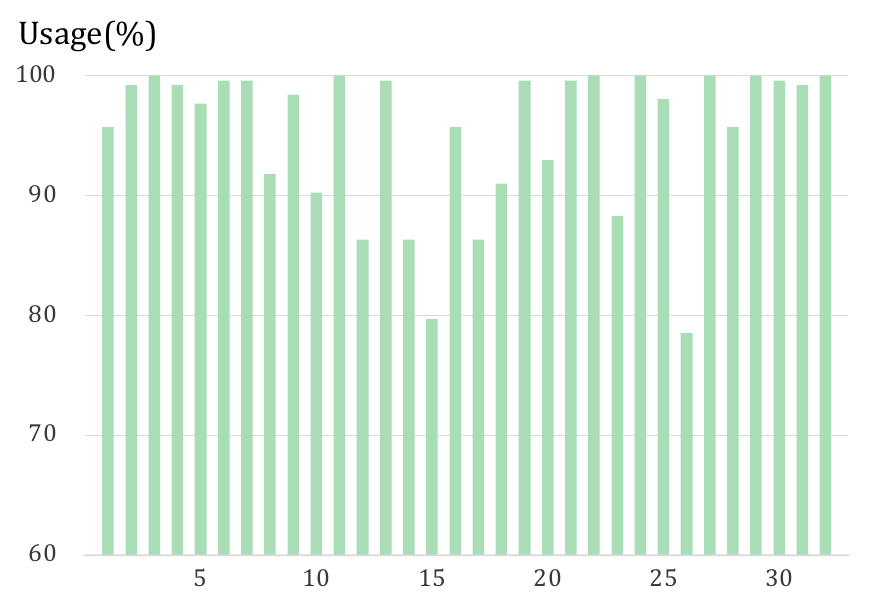}
\caption{
\textbf{Usage} of the codevectors for each segments. 
The experiment is conducted on the CIFAR10 dataset with $K=256$ and $m=32$.
A high per-segment usage rate of codevectors also suggests a considerable need for codevector sharing between different segments.
}
\label{fig_per-segment_usage}  
\end{figure}

\begin{figure}[tb]
\centering
\includegraphics[width=0.85\linewidth]{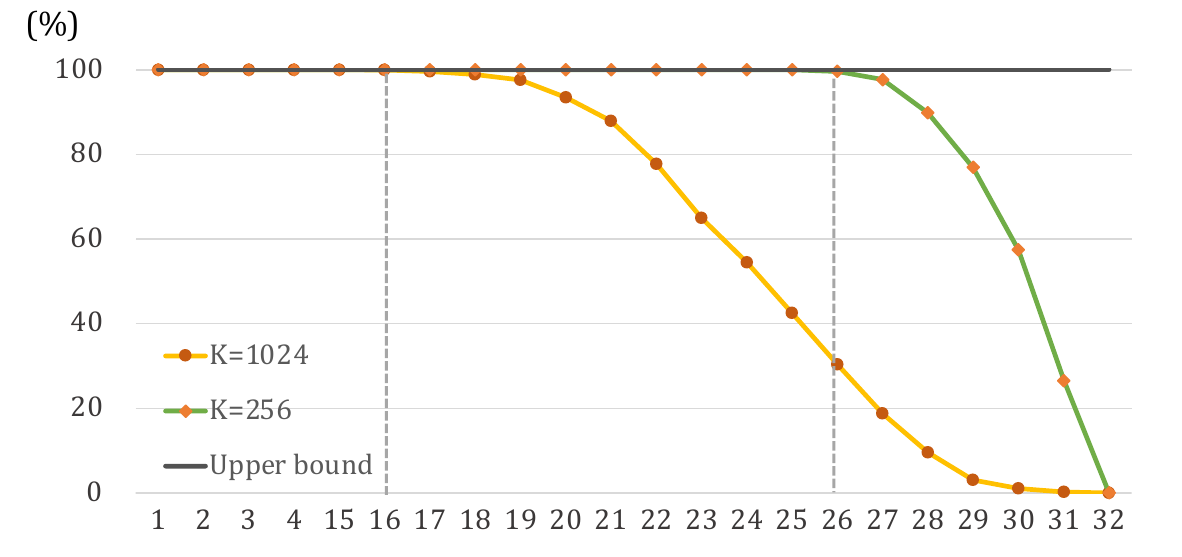}
\caption{
\textbf{(a)} 
With $K=256$, codevectors are shared among at least 26 out of 32 segments.
\textbf{(b)} 
With $K=1024$, codevectors are shared among at least 16 segments.
Smaller $K$ values result in higher codevector sharing rates.
Larger $K$ values reduce the need for a high sharing rate.
}
   \label{fig_codevector_sharing}
\end{figure}

\section{Codebook Usage}

\subsection{Usage of Codevectors}
\paragraph{Overall Usage.}
In Tab.~\ref{tab_reconstruction_results_ffhq_imagenet}, our method shows a remarkable overall usage of 100\% of the codebook. 
In this section, we employ tSNE for visualization purposes to gain insights into the learned codebooks of both VQ-VAE and our proposed LooC-VAE. 
Fig.~\ref{fig_distribution}(a) showcases the codebook visualization of VQ-VAE, where it becomes apparent that numerous codevectors remain unused. 
However, in the case of our LooC-VAE, as depicted in Fig.~\ref{fig_distribution}(b) and (c), the usage rate reaches 100\% when employing codebook sizes of 1024 and 256, respectively. This demonstrates the effectiveness of our approach in fully utilizing the available codebook capacity.

\begin{table}[b]
\centering
\begin{tabular}{lccc}
\hline
\multicolumn{1}{c}{\textbf{Method}} & \textbf{dataset} & \textbf{~$K\times d^*$} ↓~  & \textbf{usage} \\ \hline
LooC-VAE & \multirow{2}{*}{FFHQ}    & $256 \times 4$           & 100\% \\
LooC-VAE &                          & $1024 \times 4$          & 100\% \\ \hline
\end{tabular}
\caption{
\textbf{Usage of codevectors} of our LooC without CVQ's codebook update strategy.
}
\label{tab_usage_ffhq_woCVQ}
\end{table}
\paragraph{Codebook usage without CVQ’s update method}
To further investigate this, we conduct experiments by removing the CVQ update strategy in LooC and counting the codebook usage. We calculate the usage on large dataset FFHQ~\cite{karras2019ffhq}. The results are shown in Tab.~\ref{tab_usage_ffhq_woCVQ}. Our findings indicate that the codebook usage is $100\%$ at both $K = 256$ and $K = 1024$. This clearly reveals that LooC’s strength on codebook usage is not simply stemmed from the CVQ update strategy.

\paragraph{Per-segment Usage.}
As we divide the feature map into $m$ segments in the compositional VQ, we analyze each segment's codebook usage in this study.
We take CIFAR10 as the dataset in our analysis and consider the case with $K=256$ and $m=32$, as shown in Tab.~\ref{fig_reconstruction}.
When considering the codebook usage accross all 32 segments collectively, the overall usage remained 100\%. 
Fig.~\ref{fig_per-segment_usage} shows the per-segment usage of codevectors.
The horizontal axis corresponds to the segment index, while the vertical axis represents the codebook usage counted on each segment.
Upon closer examination of the individual segments, we discover that except for 6 segments, the usages on the remaining segments exceed 90\%.

\subsection{Codevector Sharing} 
In our previous usage analysis, the high codebook usage of each segment indicated significant codevector sharing among different segments. 
This higher usage is directly correlated with a higher sharing degree. 
To further explore the impact of different K values on codevector sharing, we analyze with $K=256$ and $K=1024$, using $m=32$ segments on CIFAR10 dataset. 
We analyze the percentage of codevectors that are shared by a minimum of $n \in \{1,2, \cdots,32 \}$ segments.
Fig.~\ref{fig_codevector_sharing}(a) shows that when $K=256$, all codevectors are shared among at least 26 out of 32 segments, most of which are shared among at least 30 segments. 
Fig.~\ref{fig_codevector_sharing}(b) reveals that with $K=1024$, all codevectors are shared among at least 16 segments, most of which are shared among at least 20.
A smaller $K$ value results in a higher codevector sharing rate between different segments. 
This finding explains why our method performs better with a smaller $K$. Conversely, and a larger $K$ value diminishes the urgency for pursuing a high sharing rate.

\begin{table*}[ht]
\centering
\scalebox{0.95}
{
\begin{tabular}{lcrcccccc}\hline
\textbf{Method}  &         &       $K \times d^*$~↓~  & \textbf{$l_1$ loss ↓} & \textbf{~LPIPS ↓} & \textbf{~rFID ↓} & \textbf{~SSIM ↑} & \textbf{~PSNR ↑}  \\ \hline
VQ-VAE~\cite{van2017neural_VQ-VAE} 
& \multirow{6}{*}{\rotatebox{90}{MNIST}}       & $1024 \times 128$ & 0.0207 & 0.0282  & 3.43    & 0.9777  & 26.48   \\
HVQ-VAE~\cite{williams2020hierarchical}     &  & $1024 \times 128$ & 0.0202 & 0.0270  & 3.17    & 0.9790  & 26.90   \\
SQ-VAE~\cite{takida2022_sq_vae}             &  & $1024 \times 128$ & 0.0197 & 0.0256  & 3.05    & 0.9819  & 27.49   \\
CVQ-VAE~\cite{zheng2023online_CVQ}          &  & $1024 \times 128$ & 0.0180 & 0.0222  & 1.80    & 0.9833  & 27.87   \\
\textbf{LooC$(32\times4)$ }                 &  & $32 \times ~~~~4$ & 0.0082 & \textbf{0.0083}  & \textbf{1.70}  & \textbf{0.9961}  & \textbf{35.15} \\
\textbf{LooC$(256\times4)$}                 &  & $256 \times ~~~~4$ & 0.0062 & \textbf{0.0058}  & \textbf{1.31}  & \textbf{0.9976}  & \textbf{37.58} \\ \hline
VQ-VAE~\cite{van2017neural_VQ-VAE}
& \multirow{6}{*}{\rotatebox{90}{CIFAR10}}     & $1024 \times 128$ & 0.0527 & 0.2504  & 39.67   & 0.8595  & 23.32   \\
HVQ-VAE~\cite{williams2020hierarchical}     &  & $1024 \times 128$ & 0.0533 & 0.2553  & 41.08   & 0.8553  & 23.22   \\
SQ-VAE~\cite{takida2022_sq_vae}             &  & $1024 \times 128$ & 0.0482 & 0.2333  & 37.92   & 0.8779  & 24.07   \\
CVQ-VAE~\cite{zheng2023online_CVQ}          &  & $1024 \times 128$ & 0.0448 & 0.1883  & 24.73   & 0.8978  & 24.72   \\
\textbf{LooC-VAE }                 &  & $32 \times ~~~~4$ & 0.0189 & \textbf{0.0435}  & \textbf{24.53} & \textbf{0.9805}  & \textbf{32.22} \\
\textbf{LooC-VAE}                 &  & $256 \times ~~~~4$ & 0.0144 & \textbf{0.0285}  & \textbf{19.22} & \textbf{0.9880}  & \textbf{34.51} \\ \hline
VQ-VAE~\cite{van2017neural_VQ-VAE}
& \multirow{4}{*}{\scalebox{1.1}{\rotatebox{90}{\thead{FASHION- \\ MNIST}}}}    
                                               & $1024 \times 128$ & 0.0377 & - & 12.73  & - &  23.93   \\
CVQ-VAE~\cite{zheng2023online_CVQ}          &  & $1024 \times 128$ & 0.0344 & - & 8.85  & -  & 24.66   \\
\textbf{LooC-VAE }                 &  & $32 \times ~~~~4$ &  0.0145 & 0.0158 & 8.7310 & 0.9864 & 32.4850 \\ 
\textbf{LooC-VAE}                 &  & $256 \times ~~~~4$&  0.0103 & 0.0098 & 6.2368 & 0.9924 & 35.3393 \\ \hline
\end{tabular}
} 
\caption{
\textbf{Reconstruction results} on the validation sets of MNIST~\cite{lecun1998mnist}, CIFAR10~\cite{krizhevsky2009cifar10}, and FASHION-MNIST~\cite{xiao2017fashionmnist}.
Our approach outperforms other SOTA methods and maintains comparable results even with significant reductions in the codebook size.
}
\label{tab_reconstruction_results_extension}
\end{table*}


\begin{table*}[ht]
\centering
\scalebox{1.0}
{
\begin{tabular}{lrcccccc}
\hline 
\textbf{Method} 
&  
& \textbf{$K\times d^*$} ↓~  
& \textbf{Usage↑} 
& \textbf{~LPIPS↓} 
& \textbf{~rFID↓}  
& \textbf{~SSIM↑}
& \textbf{~PSNR↑}   \\ \hline
VQGAN~\cite{esser2021taming_vq_gan} 
& \multirow{8}{*}{\rotatebox{90}{FFHQ}}     & ~~$ 1024 \times 256  $ & 42\%   & 0.1175  & 4.42 & 0.6641  & 22.24   \\
ViT-VQGAN~\cite{yu2022vector_ViT_VQGAN}   & & ~~$ 8192 \times ~~32 $ & –      & –       & 3.13 & –       & –       \\
RQ-VAE~\cite{lee2022autoregressive_RQ_VAE}& & ~~$ 2048 \times 256  $ & –      & 0.1302  & 3.88 & 0.6700  & 22.99   \\
MoVQ~\cite{zheng2022movq}                 & & ~~$ 1024 \times ~~64 $ & 56\%   & 0.0585  & 2.26 & 0.8212  & 26.72   \\
SeQ-GAN~\cite{gu2022rethinking}           & & ~~$ 1024 \times 256  $ & 100\%  & –       & 3.12 & –       & –       \\
CVQ-VAE~\cite{zheng2023online_CVQ}        & & ~~$ 1024 \times 256  $ & 100\%  & 0.0533  & 2.03 & 0.8398  & 26.87   \\
\textbf{LooC-VAE}  &    & \textbf{~~$ 256  \times ~~~~4$} & 100\% & 0.0501 & \textbf{1.97} & \textbf{0.8499} & \textbf{27.73} \\ 
\textbf{LooC-VAE}  &    & \textbf{~~$ 1024 \times ~~~~4$} & 100\% & 0.0346 & \textbf{1.37} & \textbf{0.9276} & \textbf{32.44} \\ \hline
VQGAN~\cite{esser2021taming_vq_gan}   & \multirow{8}{*}{\rotatebox{90}{ImageNet}}  
                                             & ~~$ 1024 \times 256  $ & 44\%  & 0.2011  & 7.94 & 0.5183 & 19.07      \\
ViT-VQGAN~\cite{yu2022vector_ViT_VQGAN}    & & ~~$ 8192 \times ~~32 $ & 96\%  & –       & 1.28 & –      & –     \\
RQ-VAE~\cite{lee2022autoregressive_RQ_VAE} & & $ 16384 \times 256  $ & –     & –       & 1.83 & –      & –     \\
MoVQ~\cite{zheng2022movq}                  & & ~~$ 1024 \times ~~64 $ & 63\%  & 0.1132  & 1.12 & 0.6731 & 22.42 \\
SeQ-GAN~\cite{gu2022rethinking}            & & ~~$ 1024 \times 256  $ &100\%  & –       & 1.99 & –      & –     \\
CVQ-VAE~\cite{zheng2023online_CVQ}         & & ~~$ 1024 \times 256  $ &100\%  & 0.1099  & 1.57 & 0.7115 & 23.37  \\
\textbf{LooC-VAE} &   & \textbf{~~$ 256  \times ~~~~4$}   & 100\% & 0.0916 & \textbf{1.68}   & \textbf{0.7233}  & \textbf{23.64} \\
\textbf{LooC-VAE} &   & \textbf{~~$ 1024 \times ~~~~4$}   & 100\% & 0.7160 & \textbf{1.01}   & \textbf{0.7160}  & \textbf{29.15} \\ \hline
\end{tabular}
}
\caption{\textbf{Reconstruction results} on FFHQ~\cite{karras2019ffhq} and ImageNet~\cite{deng2009imagenet}.
Our approach surpasses other SOTA methods and delivers comparable results even with significant reductions in the codebook size.
}
\label{tab_reconstruction_results_ffhq_imagenet_extension}
\end{table*}

\section{Implementation Details}
In the image reconstruction task, we compare LooC with various VQ modules. To ensure a fair comparison, we integrate LooC into VQ-VAE's network structure by replacing the VQ module, following CVQ~\cite{zheng2023online_CVQ}.
We also reimplement the quantizers in PQ~\cite{jegou2010product} and, like other approaches, apply the VQ-VAE structure for training.
Afterwards, we assess the generalizability of our LooC method on larger datasets by employing the VQ-GAN~\cite{esser2021taming_vq_gan} architecture.
Our experiments utilize the same encoder and decoder as CVQ-VAE~\cite{zheng2023online_CVQ}, both based on convolutional neural networks.
Our model was trained using a single RTX3090 GPU, and the settings for optimizer, batch size, learning rate, and number of epochs are consistent with CVQ

For image generation task, we use the LDM framework~\cite{rombach2022high} and replace VQ module with LooC and other comparative methods.
Similar to the baseline LDM, we generate our results at a resolution of $256 \times 256$ and utilize the same training parameters. 
We also adopt the same $16 \times$ downsampling scales in the latent representations as LDM, except for utilizing different quantizers.

\section{More Experimental Results}
\subsection{Quantitative Results of Reconstruction}


In Fig.~\ref{fig_teaser_extension}, we compare our results with the various latest quantizers on CIFAR10~\cite{krizhevsky2009cifar10}. We vary the number of codevectors, denoted as K, and extend Fig.~\ref{fig_teaser}-Right by including evaluation scores for four additional metrics: $l_1$ loss, LPIPS, rFID, and PSNR. 
This experiment shows that our method effectively utilizes the minimum number of codevectors to fully exploit the advantages of compositional VQ, thereby improving the effectiveness of VQ. In contrast, other state-of-the-art (SOTA) methods rely heavily on large codebooks.


Tab.~\ref{tab_reconstruction_results_extension} presents a comprehensive comparison between our method and the SOTA methods on three datasets: MNIST~\cite{lecun1998mnist}, CIFAR10~\cite{krizhevsky2009cifar10}, and FASHION-MNIST~\cite{xiao2017fashionmnist}. This table serves as an extension of Tab.~\ref{tab_reconstruction_results}. 
Our method showcases remarkable performance by achieving similar effects on rFID using a smaller codebook size of $32 \times 4$, compared to the SOTA method that requires a larger codebook of $1024 \times 128$. Notably, our method outperforms the SOTA method in terms of $l_1$ loss, LPIPS, SSIM and PSNR, indicating superior performance.
Furthermore, when our method utilizes a codebook size of $1024 \times 4$, we observe even more impressive results across various indicators.


According to the information provided, Tab.~\ref{tab_reconstruction_results_ffhq_imagenet_extension} serves as an extension of Tab.~\ref{tab_reconstruction_results_ffhq_imagenet}, with added results for the LPIPS metric on FFHQ and ImageNet datasets. The results show that our method outperforms the previous SOTA method in all indicators. Additionally, our method utilizes a smaller codebook size, specifically only one 256th of that of CVQ-VAE.

\begin{table}[bt]
\centering
\scalebox{0.8}
{
\begin{tabular}{lccccc}
\hline
\multicolumn{1}{c}{\textbf{Method}} 
& \textbf{dataset} 
& \textbf{$K\times d^*$} ↓~  
& \textbf{rFID↓}               
& \textbf{SSIM↑}                 
& \textbf{PSNR↑}                \\ \hline
PQ~\cite{jegou2010product}  
& \multirow{2}{*}{FFHQ}         & $ 16 \times 4 \times$ \#$64$  & 2.43 & 0.8054 & 25.55 \\
LooC-VAE                    &   & $1024 \times 4$               & 1.37 & 0.9276 & 32.44 \\ \hline
PQ~\cite{jegou2010product}  
& \multirow{2}{*}{ImageNet}     & $ 16 \times 4 \times$ \#$64$  & 1.96 & 0.6701 & 21.73 \\
LooC-VAE                    &   & $1024 \times 4$               & 1.01 & 0.7160 & 29.15 \\ \hline                     
\end{tabular}
}
\caption{
\textbf{Image reconstruction} results of PQ~\cite{jegou2010product}  and LooC on high-resolution datasets of FFHQ~\cite{karras2019ffhq} and ImageNet~\cite{deng2009imagenet}.
}
\label{tab_reconstruction_results_ffhq_imagenet_PQ}

\end{table}
In Tab.~\ref{tab_reconstruction_results_ffhq_imagenet_PQ}, we provide quantitative evaluation results of PQ~\cite{jegou2010product} on the FFHQ and ImageNet datasets. LooC exhibits clear superiority over PQ with regards to processing high-resolution images.
From Tab.~\ref{tab_reconstruction_results_ffhq_imagenet_PQ} and the results of other comparison methods in Tab.~\ref{tab_reconstruction_results_ffhq_imagenet_extension}, we can see that  PQ outperforms many other methods, while LooC shows clearly superior performance than PQ. Furthermore, we believe that a unified codebook is a more succinct and plausible solution.

\begin{table}[hpbt]
\centering
\begin{tabular}{llcc}
\hline
\multicolumn{1}{c}{\textbf{Method}} 
& \multicolumn{1}{c}{\textbf{dataset}} 
& $K \times d^*$~↓~ 
& \textbf{FID↓} \\ \hline
LooC-VAE                            & ImageNet-cls-avg                     & 1024 × 4          & 46.78 \\
LooC-VAE                            & LSUN-churches                        & 1024 × 4          & 15.17 \\
LooC-VAE                            & LSUN-bedroom                         & 1024 × 4          & 17.52 \\ \hline
\end{tabular}
\caption{
\textbf{Comparison of FID scores for class-conditional synthesis} on ImageNet~\cite{deng2009imagenet} and LSUN~\cite{yu15lsun}. The FID score for each class in ImageNet is computed individually and then averaged for all classes.
}
\label{tab_generation_fid}
\end{table}

\subsection{Quantitative Results of Generation}
Apart from the visualization results in Fig.~\ref{fig_generation}, we provide the comparison of FID metrics for class-conditional synthesis on ImageNet and LSUN in Tab.~\ref{tab_generation_fid}. The FID on ImageNet is 46.78, and the FIDs on LSUN-churches and LSUN-bedroom are 15.17 and 17.52 respectively. Note that the FID score for each class in ImageNet is computed individually and then averaged for all classes.


\begin{table}[t]
\centering
\begin{tabular}{lcccc}
\hline
\textbf{Method} & \textbf{~$K\times d^*$} ↓~      
& \textbf{rFID↓}               
& \textbf{SSIM↑}                 
& \textbf{PSNR↑} \\ \hline
LooC-VAE        &                           & 5.66 & 0.8141 & 26.36 \\
LooC-VAE        & \multirow{-2}{*}{256 × 4} & 3.86 & 0.8234 & 26.73 \\ \hline
\end{tabular}
\caption{
\textbf{Generalization ability}. Image reconstruction results of LooC which is trained on FFHQ~\cite{karras2019ffhq} and tested on CelebA.  %
}
\label{tab_reconstruction_results_ffhq_celeba}
\end{table}

\section{Generalization Ability}
\begin{figure*}[bt]
\centering
\includegraphics[width=0.98\linewidth]{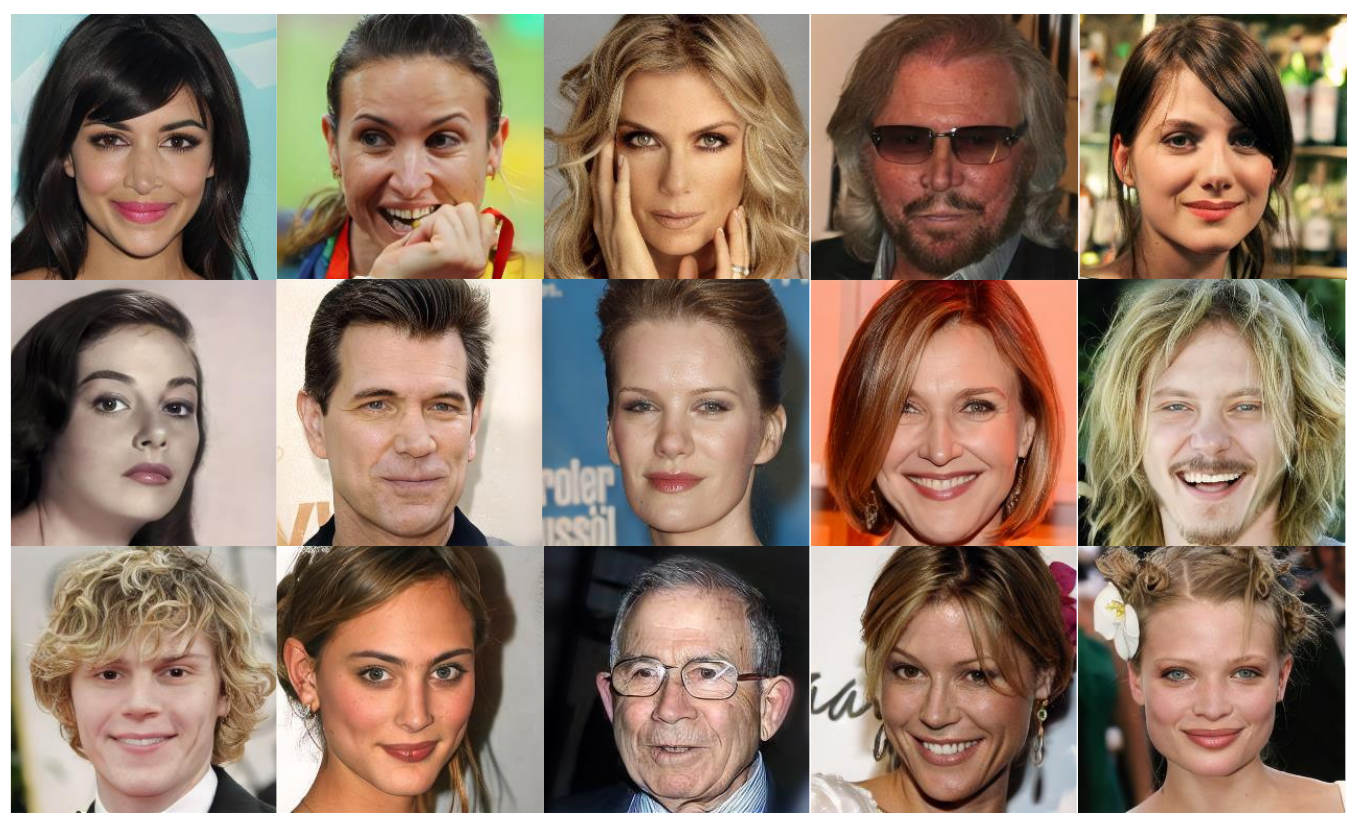}

\caption{
\textbf{Visualization of image reconstruction} of LooC, trained on FFHQ and tested on CelebA
}
\label{fig_generation_celeba}
\end{figure*}

To further investigate the generalization ability of our method, we conduct experiments by training the reconstruction model with our LooC plugged in on FFHQ and testing it on the CelebA dataset.
Tab.~\ref{tab_reconstruction_results_ffhq_celeba} and Fig.~\ref{fig_generation_celeba} showcase the quantitative and qualitative results respectively. The results demonstrate the effectiveness and strong generalization ability of our method. Despite being trained only on the FFHQ dataset, our method has achieved an rFID of 5.66 with $K = 256$ on the CelebA validation set. This is a notable improvement over VQGAN~\cite{esser2021taming_vq_gan}, which has an rFID of 10.2 with $K = 400$. Furthermore, when LooC uses $K = 1024$, the rFID is further improved to 3.86. 
The visualizations in Fig.~\ref{fig_generation_celeba} illustrate that our approach can produce intricate and high-quality reconstructions.


\section{Further Discussion}

\paragraph{Research Vision}
Current codebooks in VQ are dataset-specific, the same for LooC. A more efficient solution is a universally applicable codebook shared across different datasets and different types of data.
Exploring a cross-dataset universal codebook is a promising and valuable future research direction.
As the diversity of cross-dataset data increases, it also imposes higher requirements on the capacity of the codebook. 
Therefore, a more compact codebook design with even higher capacity remains an intriguing topic to study, LooC serves as a cornerstone for future exploration in this direction.

\paragraph{Other Impact}
The aim of this paper is to investigate a more efficient and compact method for representing visual data. Our approach helps to lower storage and transmission expenses, facilitating data exchange and sharing in our daily life. 
Meanwhile, it can expedite scientific research and foster technological innovation. 
Additionally, our approach involves dividing features into sub-segments, rather than treating them as separate and complete features. This compositional nature not only strengthens data security and privacy protection but also reduces the risk of data leakage, ultimately safeguarding the data assets of individuals and organizations.

\end{document}